\documentclass[preprint,12pt]{elsarticle}

\usepackage{amssymb}
\usepackage{amsmath}

\usepackage{algorithm}
\usepackage{algpseudocode}
\usepackage{listings}
\usepackage{lipsum}
\usepackage{mathtools}
\usepackage{booktabs}
\usepackage{tikz}
\usepackage{listings}
\usepackage{subcaption} 
\usepackage{color}
\usepackage{enumitem}
\usepackage{glossaries}
\usepackage{booktabs}
\usepackage{tabularx}
\usepackage{makecell}
\usepackage{multirow}
\usepackage{siunitx}
\usepackage{url}
\usepackage[hidelinks]{hyperref}
\usepackage{breakurl} 
\usepackage{etoolbox}
\usepackage{tikz}
\usepackage{adjustbox}
\usetikzlibrary{arrows.meta,positioning}
\usepackage{longtable}
\usepackage{ragged2e}
\usepackage{graphicx}

\newacronym{inl}{INL}{Idaho National Laboratory}
\newacronym{moose}{MOOSE}{Multiphysics Object-Oriented Simulation Environment}
\newacronym{llm}{LLM}{large language model}
\newacronym{mcp}{MCP}{Model Context Protocol}
\newacronym{rag}{RAG}{retrieval-augmented generation}
\newacronym{hit}{HIT}{Hierarchical Input Text}
\newacronym{a2a}{A2A}{Agent to Agent}
\newacronym{hpc}{HPC}{high-performance computing}
\newacronym{mms}{MMS}{Method of Manufactured Solutions}
\newacronym{vv}{V\&V}{verification and validation}
\newacronym{vtb}{VTB}{Virtual Test Bed}
\newacronym{ai}{AI}{Artificial Intelligence}
\newacronym{api}{API}{Application Programming Interface}
\newacronym{cfd}{CFD}{Computational Fluid Dynamics}

\newglossaryentry{moosenger}{
  name={MOOSEnger},
  description={MOOSEnger modeling and simulation agent framework}
}

\begin{document}

\begin{frontmatter}

\lstdefinestyle{mmsprompt}{
  basicstyle=\ttfamily\footnotesize,
  breaklines=true,
  breakatwhitespace=false,
  columns=fullflexible,
  keepspaces=true,
  showstringspaces=false,
  frame=single,
  xleftmargin=0.5em,
  xrightmargin=0.5em,
  aboveskip=0.8em,
  belowskip=0.8em
}



\title{MOOSEnger: A Simulation-Aware AI Agent Framework for the MOOSE Ecosystem}

\author[inst1]{Mengnan Li}
\ead{mengnan.li@inl.gov}
\author[inst1]{Jason Miller}
\author[inst1,inst2]{Zaid Abulawi\fnref{internship}}
\author[inst1]{Zachary Prince}
\author[inst1,inst3]{Matt Kohl\fnref{internship}}
\author[inst1]{Jack M. Cavaluzzi}
\author[inst1,inst4]{Guillaume Giudicelli}
\author[inst1]{Casey T. Icenhour}
\author[inst1]{Alexander Lindsay}
\author[inst1]{Cody Permann}

\affiliation[inst1]{
    organization={Idaho National Laboratory},
    addressline={995 MK Simpson Blvd},
    city={Idaho Falls},
    postcode={83401},
    state={ID},
    country={USA}
}

\affiliation[inst2]{
    organization={Department of Nuclear Engineering, Texas A\&M University},
    city={College Station},
    state={TX},
    country={USA}
}

\affiliation[inst3]{
    organization={University of Illinois Chicago},
    city={Chicago},
    state={IL},
    country={USA}
}

\affiliation[inst4]{
    organization={Massachusetts Institute of Technology},
    city={Cambridge},
    state={MA},
    country={USA}
}

\fntext[internship]{Work performed during internships at Idaho National Laboratory.}

\begin{abstract}
\gls{moosenger} is a modeling and simulation (M\&S) \gls{ai} agent framework for the
\gls{moose} ecosystem, built around a simulation-aware harness that combines an
interchangeable reasoning model with grounded domain knowledge, revised
simulation artifacts, MOOSE-specific validation, and executable solver
feedback. This surrounding system addresses a central limitation of one-shot
\gls{llm} generation: small syntax, schema, reference, or solver-configuration
errors can prevent a plausible input from executing, while successful execution
alone does not establish scientific correctness.

\gls{moosenger} implements a generate--check--repair--run workflow that combines
retrieval over MOOSE documentation and examples with \gls{hit}-aware parsing,
framework syntax metadata, language-server diagnostics, cross-block and
mesh-symbol checks, controlled artifact revision, and local or
\gls{mcp}-backed validation and execution. The harness binds diagnostic and
execution evidence to specific input revisions and uses that evidence to guide
bounded repair before accepting a generated artifact.

We evaluate the framework on 200 prompts spanning eight simulation families,
using both GPT~5.2 \gls{api} and Gemma~4~31B with and without the harness. For GPT~5.2,
executable success increases from 10/200 cases (5\%) in standalone generation
to 179/200 cases (89.5\%) with \gls{moosenger}. For Gemma~4~31B, success increases
from 0/200 to 153/200 cases (76.5\%). A complementary ten-case \gls{mms}
benchmark moves beyond executability: all ten generated inputs satisfy the
semantic-alignment criterion, and eight execute successfully while meeting the
prescribed single-mesh numerical-accuracy criterion.

These results show that executable reliability depends on the complete agent
system rather than on the reasoning model alone, and that simulation-aware
harnessing provides a path toward physics-informed verification and future full
application-level and engineering \gls{vv} implementation.
\end{abstract}

\begin{highlights}
\item \gls{moosenger} combines MOOSE-specific knowledge retrieval, persistent input editing, diagnostics, and solver execution in an iterative generate--check--repair--run workflow.
\item Its modular model and execution interfaces support frontier or locally deployable \gls{llm} and local or remote MOOSE applications.
\item The same harness raises executable success from 5\% to 89.5\% for the GPT~5.2 \gls{api} and from 0\% to 76.5\% for the locally deployable Gemma~4~31B on a 200-prompt benchmark.
\item A ten-case method of manufactured solutions benchmark extends assurance beyond executability, with 10/10 semantic-alignment passes and 8/10 executable numerical-accuracy passes.
\end{highlights}


\end{frontmatter}



\lstset{
    basicstyle=\ttfamily\footnotesize,
    keywordstyle=\color{green}\bfseries,
    commentstyle=\color{gray},
    stringstyle=\color{blue},
    breaklines=true,
    frame=single,
    lineskip=-1pt,
}

\glsresetall

\section{Introduction}
Frontier reasoning models are becoming increasingly adept at planning, code synthesis, and interactive problem solving. However, scientific computing exposes a limitation of model-only assistance: the desired output is not merely a paragraph or a code snippet, but an executable artifact that must satisfy domain syntax, solver constraints, mesh references, physical assumptions, and provenance requirements. A plausible-looking simulation input may still be invalid or scientifically misleading unless the surrounding system can check the artifact for simulation-specific syntax, symbols, parameters, numerical execution, and physical intent.

Harness engineering provides a useful abstraction for this surrounding system. In recent agent-system work, the harness is the runtime layer that mediates how a model observes a task, receives context, calls tools, maintains state, verifies work, attributes failures, records traces, and determines when an episode is complete~\cite{zhong2026aiharness, lin2026agenticharness}. This framing matters because agent reliability is often determined less by a single model response than by the environment, feedback channels, and verification gates wrapped around the model. Yet a general harness engineering framework is still incomplete for scientific modeling and simulation (M\&S). Generic harnesses can restrict tools, store memory, or run tests, but typically do not know whether a boundary name exists in a mesh, whether a material covers an active subdomain, whether a kernel is compatible with a given variable, whether an input file matches the target application's schema, or whether the simulation initializes successfully in the target solver.

We use the term ``simulation-aware harness engineering'' to describe the specialization required to close this gap. Rather than relying on additional prompting alone, this approach incorporates simulation semantics and authoritative framework feedback into the agent runtime. Domain knowledge guides model construction, deterministic checks assess consistency among model components, solver execution evaluates the actual candidate, and the resulting evidence is bound to the exact model revision examined. A simulation-aware harness therefore evaluates not only whether a generated model is syntactically valid and executable, but also whether it remains consistent with the intended physical problem. This distinction is essential because a simulation may execute successfully while still encoding incorrect equations, boundary conditions, material behavior, or physical assumptions.

The \gls{moose}, developed at Idaho National Laboratory, provides a representative environment for exploring this approach. MOOSE is a high-performance computing framework for multiphysics simulation~\cite{harbour2025moose}. Its modular architecture enables physics applications to be assembled from reusable components representing governing equations, boundary conditions, material behavior, solver settings, and outputs, while the framework manages the underlying parallel numerical infrastructure. This architecture has supported a broad ecosystem of MOOSE-based applications in nuclear engineering and related multiphysics domains, including reactor physics~\cite{wang2025griffin}, thermal fluids~\cite{novak2021pronghorn}, and structural mechanics~\cite{williamson2016bison}, together with shared physics modules and specialized applications~\cite{lindsay2023moose,hansel2024,icenhour2024electromagnetics,slaughter2023moose,prince2024optimization}.

Users configure MOOSE simulations through human-readable ``.i'' files written in the \gls{hit} format. A typical input file contains block-structured sections such as \texttt{[Mesh]}, \texttt{[Variables]}, \texttt{[Kernels]}, \texttt{[BCs]}, \texttt{[Materials]}, and \texttt{[Executioner]}, with nested sub-blocks and parameters that instantiate objects registered by a particular MOOSE-based application. This design offers substantial flexibility, but it also requires users to translate physical intent into valid \gls{hit} syntax and the application-specific object hierarchy of the target executable. Failures can occur at multiple levels, including malformed syntax, invalid object or parameter names, missing required parameters, inconsistent mesh or symbol references, incomplete material coverage, unsuitable solver settings, and modeling choices that are executable but physically inappropriate.

\gls{moosenger} provides a concrete realization of simulation-aware harness engineering for scientific modeling and simulation, with MOOSE as its computational environment. Rather than treating input authoring as isolated text generation, it composes an interchangeable reasoning model with a domain knowledge substrate, a tool-enabled agent runtime, persistent workspace state, and executable feedback from MOOSE. The system supports both closed models accessed through hosted services and open-weight models deployed locally or on self-hosted infrastructure.

Simulation awareness is introduced through MOOSE-specific grounding, authoring, analysis, and execution capabilities. \gls{rag} and syntax metadata guide model construction; \gls{hit}-aware parsing and transactional block operations support structured authoring; symbol-reference analysis, language-server diagnostics, and mesh-symbol extraction expose inconsistencies; and local or \gls{mcp}-backed execution provides direct solver feedback. Evidence from these operations is bound to the exact input revision examined, enabling acceptance gates to assess structural validity, cross-reference consistency, schema conformance, and executability together. Although these gates do not by themselves constitute complete engineering \gls{vv}, they provide an extensible assurance infrastructure into which stronger, application-specific criteria can be incorporated. The architecture can also accommodate future domain foundation models as generators, planners, rerankers, diagnostic classifiers, or repair modules without changing the assurance requirements applied to their outputs. In this way, \gls{moosenger} serves both as a practical MOOSE agent and as a platform for investigating progressively more complete engineering \gls{vv} workflows.

The main contributions of this paper are:
\begin{itemize}
\item a simulation-aware harness engineering framing that identifies the domain semantics, executable feedback, revision-bound evidence, and assurance mechanisms required beyond those provided by generic agent harnesses;

\item a model- and deployment-portable architecture that integrates hosted or locally deployed reasoning models with grounded MOOSE knowledge, revisioned simulation inputs, domain tools, validation policies, and solver feedback;

\item a MOOSE-specific realization of this architecture through \gls{hit}-aware input construction and repair, object and parameter validation, symbol and mesh consistency analysis, and local or remote executable verification;

\item an eight-family, 200-prompt controlled evaluation that isolates the effect of the harness across two reasoning models, increasing runnable-input success from \(5\%\) to \(90\%\) for GPT~5.2 \gls{api} and from \(0\%\) to \(76.5\%\) for Gemma~4~31B; and

\item two complementary higher-assurance evaluations: a ten-case \gls{mms} benchmark that separates semantic alignment from numerical accuracy, and a reference-grounded \gls{vtb} duct-bowing demonstration of multi-step model construction, diagnosis, repair, and validation.
\end{itemize}

\section{Related Work}
\label{sec:related-work}

\gls{rag} supplies language models with external evidence,
while approaches such as ReAct interleave reasoning, tool use, and environmental
observations~\cite{lewis2020rag,yao2023react}. Coding agents extend this pattern
through persistent files, command execution, and test feedback. Scientific
simulation imposes additional requirements because the generated artifact must
conform to a domain-specific input language, remain internally consistent,
execute with the target solver, and preserve the modeling intent expressed by
the user. Successful execution alone is therefore a necessary but insufficient
criterion for trustworthy simulation authoring.

Recent simulation agents increasingly close the loop between generation and
solver feedback. MetaOpenFOAM combines multi-agent task decomposition with
retrieval over OpenFOAM examples~\cite{chen2024metaopenfoam}. OpenFOAMGPT
implements an execution-feedback correction loop in which solver errors are
appended to the model context and the case is regenerated until execution
succeeds~\cite{pandey2025openfoamgpt}. This establishes the value of placing the
solver in the loop, but the reported mechanism remains comparatively
lightweight: execution failure is primarily returned as textual feedback for
another generation round, rather than being decomposed into heterogeneous
validation evidence with explicit artifact and repair-state semantics.

Foam-Agent advances solver-feedback repair through a multi-agent workflow that combines dependency-aware generation with a trajectory-conditioned Reviewer~\cite{yue2025foam}. When execution fails, the Runner Agent extracts standard OpenFOAM errors, while the Reviewer Agent diagnoses them, plans and
applies targeted multi-file changes, and conditions subsequent attempts on the history of previous errors and corrections. Its recovery actions can modify
boundary conditions, numerical schemes, time-step controls, meshes, or solver and model choices, although some runner-level and floating-point failures fall outside its error-extraction interface. The loop continues until execution succeeds or an iteration limit is reached. Foam-Agent therefore distributes recovery across specialized agents and relies primarily on solver-log evidence, whereas \gls{moosenger} centers a single interchangeable reasoning agent within a simulation-aware harness that binds multiple forms of evidence to explicit artifact revisions and controls whether to apply targeted repair, gather additional evidence, replan, promote a candidate, or terminate.

ChatCFD employs a modular multi-agent framework to automate the \gls{cfd} workflow
from user intent through case generation, execution, error correction, and
postprocessing~\cite{fan2026chatcfd}. It also evaluates the physical fidelity
of runnable cases, though this assessment remains an evaluation protocol
rather than an integrated \gls{vv} gate. Its recovery process combines
specialized error-localization and correction modules with iterative reflection,
but the reported results show diminishing returns from later reflection attempts
on persistent errors, motivating stronger knowledge integration or recovery
strategies beyond repeated reflection. \gls{moosenger} addresses these challenges
through a different architecture: a harness-controlled workflow in which structured domain knowledge and revision-bound validation evidence govern repair decisions. Its hybrid grounding
architecture combines \gls{rag}, deterministic MOOSE metadata, a \gls{llm} Wiki--inspired compiled knowledge layer, and governed promotion of derived guidance.

LARA-HPC establishes pre-execution validation as a reliability strategy for computationally expensive HPC workflows by applying syntax, APIs, physical-consistency, and simulation-native dry-run checks prior to production submission~\cite{dawson2026lara}. Structured feedback supports iterative refinement, while a controlled execution layer mediates access to HPC resources. \gls{moosenger} applies a related principle to MOOSE input authoring but emphasizes artifact governance: the reasoning model may repeatedly revise an input, while the harness maintains revision identity and controls validation and acceptance.

Within the MOOSE ecosystem, MooseAgent combines task decomposition, retrieval
over MOOSE documentation and examples, and multi-round iterative verification
of generated inputs~\cite{zhang2025mooseagent}. It demonstrates that
agent-mediated correction can improve MOOSE input generation, but the published workflow does not define a managed revision lifecycle in which validation evidence is explicitly bound to an inspected artifact revision and in which acceptance is controlled independently of the reasoning model.

AutoMOOSE addresses a complementary MOOSE problem through a specialized
multi-agent phase-field workflow~\cite{manna2026automoose}. Its agents
coordinate execution screening, physics-specific assessment, and recovery, with corrected simulations being rerun and independently re-evaluated. The workflow operates on physics-specific input templates whose plugins supply the falsification criteria. Its repair loop primarily adjusts bounded numerical and discretization parameters, rather than modifying physical parameters or reconstructing the input model. AutoMOOSE therefore provides physics-aware assessment and recovery within prescribed template-based workflows, but does not generally repair arbitrary MOOSE input structures, object selections, block hierarchies, parameter references, or formulation choices.

These studies establish complementary strategies for knowledge grounding, validation, and failure recovery. Building on these directions, \gls{moosenger} contributes a model-agnostic, simulation-aware harness that integrates domain grounding and executable validation with MOOSE-specific input-generation, diagnosis, and repair strategies for general-purpose input authoring. At its core is a control pattern that we term \emph{artifact-governed repair}: the reasoning model proposes revisions, while the harness maintains artifact identity, binds validation evidence to the exact revision inspected, and controls acceptance and termination. This separation prevents model-generated changes from being treated as validated artifacts and supports multiple evidence sources and bounded recovery. Physics-specific assessments, including those demonstrated by AutoMOOSE and ChatCFD, can be incorporated as additional acceptance criteria within this control structure. Section~\ref{sec:architecture} describes the mechanism in detail.

\section{\gls{moosenger}: Simulation-Aware Agent Architecture}
\label{sec:architecture}

\gls{moosenger} treats input generation as a controlled artifact lifecycle rather than a single model response. The reasoning model interprets intent and proposes a formulation, and the harness supplies domain evidence, controls which tools are available, owns the mutable and accepted artifacts, and decides whether the evidence is sufficient to advance the workflow. This separation creates a clear trust boundary: the model output is a proposal, whereas parser, schema, diagnostic, and solver results are observations about a specific input revision.

We refer to this control pattern as \emph{artifact-governed repair}. The model may inspect evidence and propose changes to a mutable draft, but a model response does not itself establish a new accepted simulation artifact. Each authoring operation creates an explicit draft revision. Validation evidence is associated with the exact revision inspected, and a passing revision is frozen as an immutable candidate before it can advance through the configured acceptance gates. Only the harness may promote a candidate to the input used for subsequent execution. Conversely, failed validation does not overwrite the last accepted artifact: it produces structured, revision-bound evidence usable to construct the next draft revision. Artifact-governed repair therefore separates model-driven diagnosis and revision from harness-owned validation, acceptance, promotion, and termination.

\subsection{Design Requirements and System Composition}
\label{sec:core-architecture}

Table~\ref{tab:design-requirements} summarizes the requirements that motivated the architecture. They define the controls needed before physics-specific \gls{vv} can be layered on top of an agent-generated simulation.

\begin{table}[htbp]
\caption{Design requirements and corresponding MOOSEnger mechanisms.}
\label{tab:design-requirements}
\centering
\footnotesize
\renewcommand{\arraystretch}{1.16}
\setlength{\tabcolsep}{6pt}

\begin{tabularx}{\linewidth}{
    @{}
    >{\raggedright\arraybackslash}p{0.22\linewidth}
    >{\raggedright\arraybackslash}X
    @{}
}
\toprule
\textbf{Design requirement} &
\textbf{MOOSEnger mechanism} \\
\midrule

\textbf{Domain grounding} &
Retrieval over documentation and examples, MOOSE-native object and parameter
metadata, and task-local artifacts \\
\addlinespace[3pt]

\textbf{Artifact identity} &
Revisioned drafts, immutable validation candidates, and a single accepted
input \\
\addlinespace[3pt]

\textbf{Repair authority} &
The model proposes revisions, and the harness binds evidence to the inspected
revision and controls candidate acceptance and promotion \\
\addlinespace[3pt]

\textbf{Deterministic validation} &
\gls{hit} structure, block and object schemas, and parameter, cross-link, symbol, reference, and language-server checks \\
\addlinespace[3pt]

\textbf{Executable evidence} &
Authoritative validation by the target MOOSE application, bounded probes, mesh generation, and full simulation execution \\
\addlinespace[3pt]

\textbf{Recovery control} &
Structured diagnostics, failure-progress checks, finite repair budgets, and
controlled replanning, escalation, or termination \\
\addlinespace[3pt]

\textbf{Traceability} &
Revision-bound records of validation, execution, logs, timing, and produced
artifacts \\

\bottomrule
\end{tabularx}
\end{table}

\begin{figure}[htbp]
\centering
\includegraphics[width=0.96\linewidth]{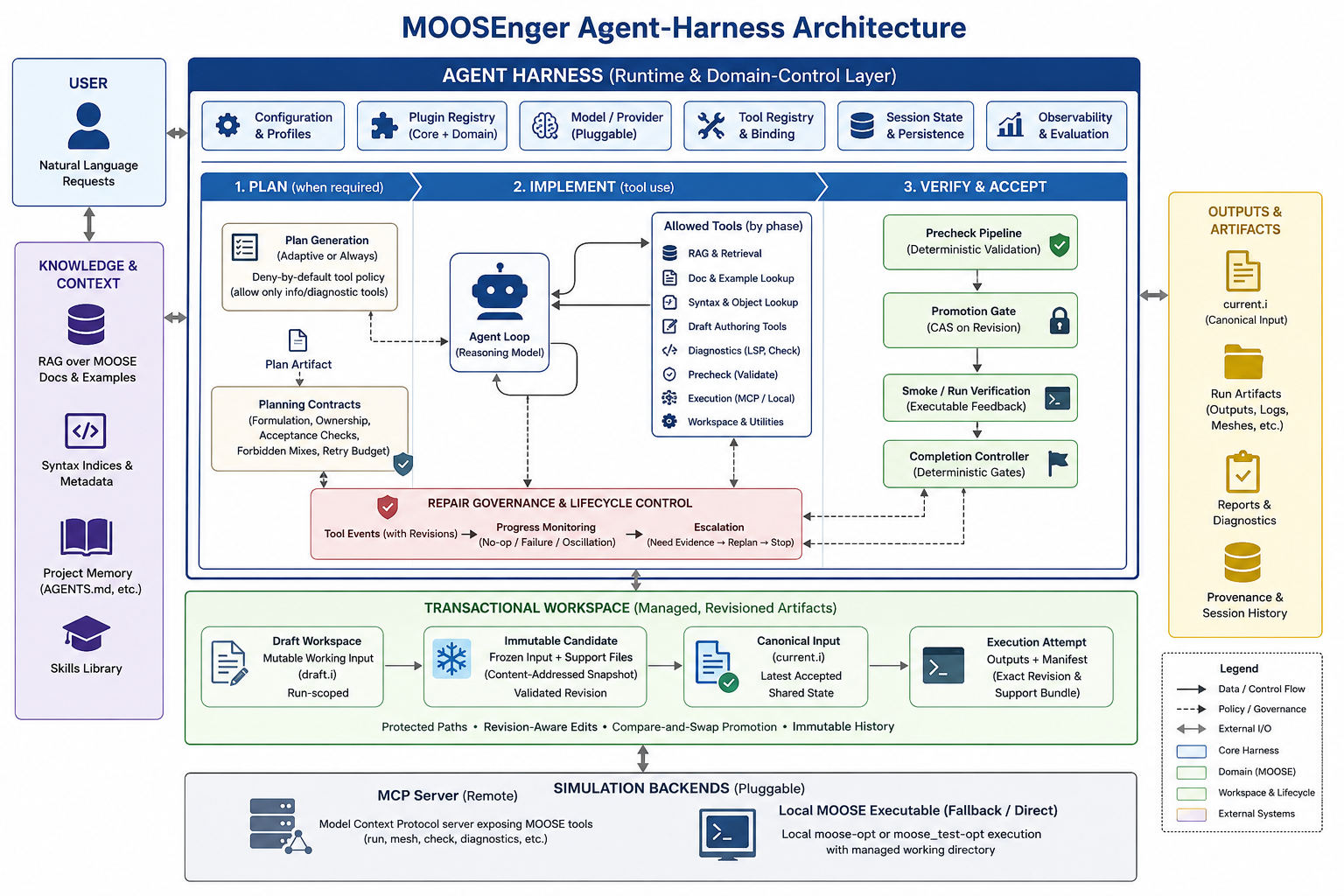}
\caption{Core-plus-domain architecture. The reasoning model is replaceable,
while the MOOSE plugin and core runtime retain the domain evidence, artifact
lifecycle, validation policy, and execution gates. Model substitution therefore changes how a candidate is produced, not the artifact-level conditions used to accept it.}
\label{fig:harness-architecture}
\end{figure}

Figure~\ref{fig:harness-architecture} shows the core-plus-domain composition. The reusable core provides configuration, provider selection, plugin and tool registries, guarded filesystem access, session state, memory and skill wiring, planning policy, and evaluation. A domain plugin supplies prompt packs, configuration schemas, importers, tools, lifecycle services, and acceptance policies. The MOOSE plugin adds \texttt{pyhit}-based parsing, syntax-aware ingestion, object and parameter lookup, managed input authoring, validation, language-server integration, and local or remote execution. Agent profiles select the prompt pack and permitted tools for a run, while a shared tool context binds configuration, workspace state, retrieval resources, and the execution backend.

The reasoning model can be accessed through a hosted provider or a locally
served interface. Model instructions and confidence are not used as acceptance evidence. Instead, tool wrappers validate inputs, enforce phase-specific access, serialize results, and register produced artifacts. During planning, the policy can restrict the agent to non-mutating evidence tools; authoring and execution tools are enabled only after the workflow advances to implementation. Optional planning contracts capture fixed user requirements separately from replaceable formulation choices, allowing a failing numerical approach to be reconsidered without discarding the original task constraints.

\subsection{MOOSE Knowledge Grounding}
\label{sec:moose-aware-ingestion}

Reliable simulation assistance requires more than a single semantic retrieval index. \gls{moosenger} therefore organizes domain knowledge as a layered, provenance-aware resource that combines MOOSE documentation and examples, structured framework metadata, compiled modeling guidance, explicit relationships among domain concepts, and governed knowledge distilled from validated cases, diagnostics, and successful repair workflows. The vector database indexes relevant evidence for retrieval but is not treated as the sole authority on framework behavior. This design is inspired in part by Karpathy's \gls{llm} Wiki pattern, in which a persistent, incrementally updated knowledge layer mediates between source evidence and runtime reasoning ~\cite{karpathy2026llmwiki}. \gls{moosenger} adapts this idea to scientific simulation by combining persistent compiled knowledge with deterministic MOOSE metadata, provenance and freshness tracking, governed knowledge promotion, and executable verification. Figure~\ref{fig:moose-knowledge-grounding} summarizes the resulting workflow.

Knowledge sources include the MOOSE repository and documentation, example inputs, machine-readable syntax metadata, application-specific documents, community discussions, and project knowledge. MOOSE-based applications can contribute documentation and examples through application knowledge packs without modifying the common knowledge base. Each source retains provenance and dependency information, while incremental synchronization updates only affected resources. Changes to authoritative syntax sources can rebuild the object registry, and derived knowledge can be marked stale when its supporting evidence changes.

MOOSE input files receive specialized treatment because a ``.i'' file is a hierarchical \gls{hit} artifact rather than ordinary prose. \gls{moosenger} uses \texttt{pyhit} to retain the complete input while creating block-aligned retrieval units for sections such as \texttt{[Mesh]}, \texttt{[Kernels]}, \texttt{[BCs]}, and \texttt{[Materials]}. These units retain their hierarchy, object types, parameters, and relationship to the parent input. A complementary registry derived from MOOSE's machine-readable syntax and documentation records systems, registered objects, parameters, descriptions, and documentation links. Deterministic parser and framework metadata remain distinct from generated summaries or topic annotations.

The compiled knowledge layer captures reusable modeling patterns, validated cases, common failure modes, and lessons from successful workflows. Newly derived guidance is represented as a structured candidate artifact containing its claim, supporting evidence, provenance, source dependencies, and relevant inputs, diagnostics, or validation results. Structural, provenance, freshness, and optional agent-level checks determine whether the artifact becomes retrieval-visible. Workflow-derived guidance may additionally require successful MOOSE validation or execution, and rejected or stale artifacts remain isolated from the active knowledge base.

Although this promotion mechanism is implemented, it currently governs
candidate artifacts only after they have been created, and is not yet routinely used for learning across independent runs. Such use also requires persistent memory that can accumulate diagnostics, successful repairs, project facts, and user preferences and transform them into promotion candidates. Preliminary work has explored this capability, but its integration with the governed promotion pipeline, as well as its systematic evaluation, remains ongoing~\cite{abulawi2026deployingfrontieragentictechnology}.

Explicit relationships connect MOOSE objects with their systems and parameters, input blocks with instantiated objects, examples with modeling patterns, and recurring failures with relevant guidance. At runtime, a unified grounding process retrieves relevant documentation, examples, framework metadata, compiled knowledge, relationships, and task-local artifacts, all while retaining source attribution. This knowledge guides formulation, input authoring, and failure interpretation but does not establish simulation validity. The simulation-aware harness remains responsible for evaluating the resulting artifact through diagnostics, validation checks, and, when required, execution of the target MOOSE application.

\begin{figure*}[t]
\centering

\begin{adjustbox}{max width=\textwidth}
\begin{tikzpicture}[
    font=\sffamily,
    >=Latex,
    stage/.style={
        rounded corners=3pt,
        very thick,
        minimum width=3.25cm,
        minimum height=0.85cm,
        align=center,
        font=\bfseries\large
    },
    panel/.style={
        rounded corners=3pt,
        draw=gray!45,
        line width=0.8pt,
        minimum width=3.25cm,
        minimum height=3.25cm,
        text width=2.85cm,
        align=center,
        inner sep=6pt,
        font=\normalsize
    },
    arrow/.style={
        ->,
        line width=1.4pt,
        draw=gray!65
    }
]

\definecolor{mooseblue}{HTML}{0F4D92}
\definecolor{moosegold}{HTML}{B68122}
\definecolor{moosegreen}{HTML}{4D8659}
\definecolor{mooseteal}{HTML}{347D80}


\node[
    stage,
    draw=mooseblue,
    fill=mooseblue!6
] (s1) at (0,0)
{1. Evidence Sources};

\node[
    panel,
    below=4mm of s1
] (p1)
{
    \textbf{MOOSE Documentation \& Examples}\\[4mm]

    \textbf{MOOSE Syntax Metadata}\\[4mm]

    \textbf{Community Discussions}\\[4mm]

    \textbf{Application Knowledge Packs}
};


\node[
    stage,
    draw=moosegold,
    fill=moosegold!6,
    right=12mm of s1
] (s2)
{2. Knowledge Build};

\node[
    panel,
    below=4mm of s2
] (p2)
{
    \textbf{Sync \& Change Detection}\\[5mm]

    \textbf{Parse \& Structure}\\[5mm]

    \textbf{Index, Compile \& Promote}
};


\node[
    stage,
    draw=moosegreen,
    fill=moosegreen!6,
    right=12mm of s2
] (s3)
{3. Grounding Layer};

\node[
    panel,
    below=4mm of s3
] (p3)
{
    \textbf{Vector RAG}\\[4mm]

    \textbf{MOOSE Object Registry}\\[4mm]

    \textbf{Promoted Domain Knowledge}\\[4mm]

    \textbf{Typed Relationships (optional)}
};


\node[
    stage,
    draw=mooseteal,
    fill=mooseteal!6,
    right=12mm of s3
] (s4)
{4. Runtime \& Verification};

\node[
    panel,
    below=4mm of s4
] (p4)
{
    \textbf{Merged Retrieval \& Attribution}\\[5mm]

    \textbf{LLM Authoring \& Repair}\\[5mm]

    \textbf{MOOSE Validation \& Execution}
};


\draw[arrow] (s1.east) -- (s2.west);
\draw[arrow] (s2.east) -- (s3.west);
\draw[arrow] (s3.east) -- (s4.west);

\end{tikzpicture}
\end{adjustbox}

\caption{
\gls{moosenger} knowledge-grounding workflow.
MOOSE documentation and examples, machine-readable syntax metadata,
community or project knowledge, and application-specific knowledge packs
are synchronized and transformed into provenance-aware knowledge resources. The grounding layer combines vector retrieval with a deterministic MOOSE object registry, promoted domain knowledge, and optional typed relationships. At runtime, relevant evidence is merged with source attribution to support input authoring and repair, while MOOSE validation and execution remain the authority for determining whether the generated simulation artifact is acceptable.}
\label{fig:moose-knowledge-grounding}

\end{figure*}

\subsection{Artifact-Governed Input Authoring, Validation, and Repair}
\label{sec:input-generation-validation}

\gls{moosenger} treats input authoring as a controlled artifact lifecycle rather than as repeated editing until the reasoning model declares success. After interpreting the request and gathering relevant evidence, the model authors a mutable draft at revision \(r\). Before the draft can support an acceptance claim, the harness freezes it---together with its supporting files---as an immutable candidate \(C_r\). All subsequent validation and execution evidence is bound to this exact candidate, maintaining a clear distinction between the artifact proposed by the model and the evidence produced by the harness.

\begin{figure}[!t]
    \centering
    \includegraphics[
        width=0.96\linewidth,
        keepaspectratio
    ]{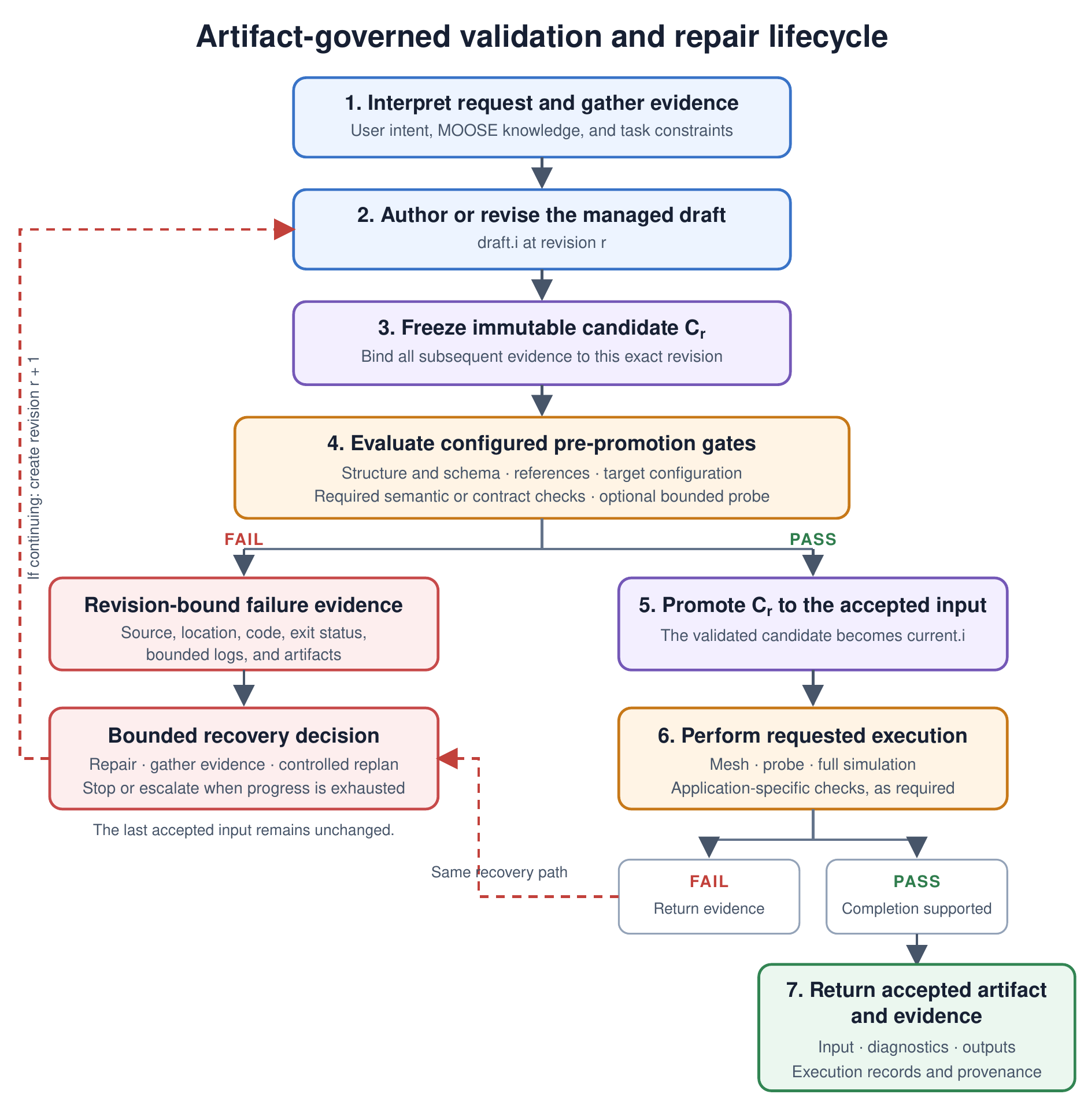}
    \caption{Artifact-governed validation and repair lifecycle. Each mutable
    draft is frozen as an immutable candidate before the configured
    pre-promotion gates are evaluated. Failed candidates produce
    revision-bound evidence for bounded repair, controlled replanning, or
    termination, while only a passing candidate may be promoted. Requested
    execution and application-specific checks provide additional evidence and
    determine whether the accepted revision can support a task-level completion
    claim.}
    \label{fig:artifact-lifecycle}
\end{figure}

Figure~\ref{fig:artifact-lifecycle} summarizes this lifecycle. The configured pre-promotion gates may combine structural and schema checks, cross-block and mesh-symbol reference validation, authoritative MOOSE configuration validation, requirement or semantic criteria, and a bounded execution probe when required. Findings from these sources are normalized into structured evidence that identifies the source, location, severity, diagnostic code, and supporting logs or artifacts. This allows the model to distinguish, for example, a malformed input structure from an invalid reference, an unsupported parameter, or a solver-configuration failure.

If a pre-promotion gate fails, \(C_r\) remains rejected and the last accepted input is left unchanged. The resulting diagnostics and artifacts are returned to the model as revision-bound repair evidence. Any correction creates a new draft revision \(r+1\), which must be frozen and evaluated independently; evidence from \(C_r\) cannot be used to certify the revised artifact. Depending on the failure, the workflow may apply a targeted repair, gather additional domain evidence, reconsider a replaceable formulation choice, or escalate to controlled replanning.

\gls{moosenger} also treats repair progress as workflow state. Repeated
diagnostics, ineffective edits, and revision oscillation are detected so that the same failure is not presented indefinitely as new evidence. Repair continues only within the configured budget. When the failure state does not improve, the harness may terminate or escalate with the unresolved evidence exposed, rather than allowing the model to continue editing without measurable progress.

When all configured pre-promotion gates pass, the harness may promote \(C_r\) to the accepted input, \texttt{current.i}. Requested mesh generation, bounded probes, full simulation execution, or application-specific checks can then produce additional evidence for that accepted revision. For tasks requiring these operations, promotion alone is insufficient for completion: the corresponding execution or application-level gate must also pass. A failure at this stage returns evidence to the same bounded recovery process without overwriting the accepted artifact or attributing the result to another revision. The lifecycle therefore separates model-driven authoring and repair from harness-controlled inspection, promotion, execution, and completion. The claims supported by the different evidence gates are summarized in Table~\ref{tab:assurance-gates}.

\subsection{Execution Evidence, Completion, and Scope of Assurance}
\label{sec:moose-execution-loop}

\gls{moosenger} exposes local and \gls{mcp}-backed execution through a common
managed runner. The local backend supports workstation or offline use, while
the \gls{mcp} backend can place execution near a configured MOOSE installation or
remote computing resource. A transport failure may trigger a configured
backend fallback, but a scientific failure returned by a reachable backend is
preserved as authoritative evidence.

The completion controller evaluates the evidence required by the user
request. An input-authoring task requires a promoted candidate, a mesh task
requires successful mesh-generation evidence, and a simulation task requires
successful execution of the accepted revision. When the request includes an
application-specific verification criterion, that criterion must also pass
before the workflow can claim completion. A fluent model response therefore
cannot substitute for missing artifact or execution evidence.

Table~\ref{tab:assurance-gates} summarizes the evidence gates and the claims
that each can support. The gates are cumulative only when required by the task:
passing a lower-level gate does not imply satisfaction of a stronger one.

\begin{table}[t]
\caption{Evidence gates and the assurance claim supported by each gate.}
\label{tab:assurance-gates}
\centering
\footnotesize
\renewcommand{\arraystretch}{1.16}
\setlength{\tabcolsep}{5pt}

\begin{tabularx}{\linewidth}{
    @{}
    >{\raggedright\arraybackslash}p{0.19\linewidth}
    >{\raggedright\arraybackslash}p{0.34\linewidth}
    >{\raggedright\arraybackslash}X
    @{}
}
\toprule
\textbf{Assurance gate} &
\textbf{Evidence examined} &
\textbf{Supported claim} \\
\midrule

\textbf{Structural validity} &
HIT structure; object and parameter schemas; symbol, reference, and
language-server findings &
Internally consistent under the implemented syntax, schema, and reference
checks. \\
\addlinespace[3pt]

\textbf{Requirement alignment} &
Explicit task constraints and their assessed representations in the candidate &
Represents the assessed user requirements within the declared evaluation
scope. \\
\addlinespace[3pt]

\textbf{Configuration validity} &
Authoritative validation by the target MOOSE application &
Accepted by the target executable at the configuration stage. \\
\addlinespace[3pt]

\textbf{Bounded execution} &
Harness-controlled execution within prescribed limits &
Initializes and advances within the configured probe limits. \\
\addlinespace[3pt]

\textbf{Requested operation} &
Exit status, logs, outputs, and artifact manifest for the requested mesh,
probe, or simulation &
Successfully completes the requested operation using the accepted revision. \\
\addlinespace[3pt]

\textbf{Application-specific verification} &
Manufactured-solution results or other explicitly defined application evidence &
Satisfies the specified numerical or \gls{vv} criterion within its defined scope. \\

\bottomrule
\end{tabularx}
\end{table}

Validation and execution records retain the evaluated candidate revision,
execution purpose, backend, effective configuration, completion status,
bounded logs, timing, and produced artifacts. This provenance is important
when multiple drafts, validation candidates, probes, and simulation runs coexist
during repair. It also provides a model-independent evaluation boundary:
different reasoning models may follow different repair trajectories, but the
final artifacts are assessed against the same configured gates.

These gates support progressively stronger but bounded claims. Structural
validity does not establish executability, successful execution does not by
itself establish semantic or physical correctness, and an
application-specific check supports only the criterion that was explicitly
evaluated. The gate structure is intentionally extensible: additional
domain- or application-specific assessments can be incorporated without
changing the underlying artifact-governed repair mechanism, provided that
their evidence is bound to the evaluated revision and their acceptance
criteria are explicitly defined. Physics-specific checks for geometry,
material data, constitutive assumptions, boundary conditions, multiphysics
couplings, conservation, convergence, or other \gls{vv} criteria can therefore be
layered onto the same control structure as required by the application. Such
extensions strengthen the assurance available for a particular workflow, but
their claims remain bounded by the criteria actually evaluated and do not
replace qualified engineering review.

\section{Interfaces and Deployment}
\label{sec:interfaces}

The simulation-aware harness described in Section~\ref{sec:architecture} is independent of the interface through which a request arrives. Engineers, evaluation pipelines, development environments, and other agents can therefore use the same MOOSE knowledge, persistent
artifacts, validation workflow, and execution backends. Interfaces determine
how requests and results are exchanged, but do not change the
artifact-level acceptance gates in Table~\ref{tab:assurance-gates}. The model
outputs remain proposals, validation evidence remains bound to specific
revisions, and only the harness can promote an accepted input.

\gls{moosenger} currently provides four access surfaces, summarized in
Table~\ref{tab:interfaces}. The command-line interface and evaluation harness
support direct interactive and batch use. The \gls{mcp}
server exposes MOOSE validation, mesh generation, and execution as callable
tools, and can place execution near a local or remote MOOSE installation. The
Agent-to-Agent (A2A) server allows other software agents to delegate complete
MOOSE tasks, while the Claude Code plugin provides an interactive
assistant-oriented interface built on the same A2A service.

\begin{table}[t]
\centering
\caption{MOOSEnger access surfaces. Each surface uses the same
artifact-governed harness and acceptance gates.}
\label{tab:interfaces}
\small
\begin{tabularx}{\linewidth}{
    @{}
    p{0.22\linewidth}
    p{0.25\linewidth}
    X
    @{}
}
\toprule
\textbf{Surface} &
\textbf{Primary use} &
\textbf{Capability} \\
\midrule

Command-line interface and evaluation harness &
Interactive and batch workflows &
Direct access to MOOSE question answering, input authoring, validation, and
execution. \\

MCP server &
Tool-enabled editors and agents &
Callable MOOSE validation, mesh generation, and execution services, including
local or remote execution. \\

A2A server &
Other software agents &
Discoverable delegation of complete MOOSE tasks with streamed progress and
optional session continuity. \\

Claude Code plugin &
Assistant and integrated development environment users &
Interactive MOOSE authoring, revision, validation, and execution within a
persistent conversation. \\

\bottomrule
\end{tabularx}
\end{table}

\subsection{Agent-to-Agent Server}
\label{sec:a2a}

The A2A server makes \gls{moosenger} available as a discoverable scientific
agent. Other agents can delegate natural-language requests for MOOSE question
answering, input generation/modification, validation, and execution--without
reproducing MOOSE-specific reasoning or tool integration in the calling
system. A machine-readable agent description advertises these capabilities so
callers can select the appropriate task programmatically.

Because simulation workflows may involve multiple authoring, validation,
repair, and execution steps, the server can stream intermediate progress and
produced artifacts before returning the final response. This makes a
long-running workflow observable to the calling agent and returns both a
natural-language summary and the generated simulation artifacts. Requests may
be handled independently or associated with a durable session, allowing
follow-up requests to revise the same workspace and continue from prior
simulation state. Whether deployed locally or as a shared service, the A2A
interface retains the same artifact-governed repair and acceptance process
as described in Section~\ref{sec:architecture}.

\subsection{Claude Code Plugin}
\label{sec:plugin}

The Claude Code plugin provides an initial demonstration of how
\gls{moosenger} can be embedded within a widely used interactive coding
environment rather than requiring users to move between a general-purpose
assistant and a separate simulation interface. From the same conversation,
users can submit MOOSE questions, request new or modified input files, invoke
validation, and launch simulation runs through a managed A2A service. Generated inputs, execution results, diagnostics, and other artifacts are returned directly to the conversation, allowing model development and inspection to continue without changing chat windows or disrupting the surrounding coding workflow.

Although the interaction occurs through Claude Code, the simulation-specific
work remains delegated to \gls{moosenger}. Each conversation is associated with a durable \gls{moosenger} session and project workspace, allowing follow-up requests to refer to earlier decisions, modify the same evolving input artifact, and retain validated state across turns or service restarts. Behind the conversational interface, \gls{moosenger} can access its MOOSE knowledge resources, apply the same artifact revision and promotion rules described above, execute structural and application-level validation gates, and invoke a configured local or remote MOOSE executable. The plugin therefore provides a familiar front end while preserving the simulation-aware grounding, validation, execution, and provenance mechanisms of the underlying harness.

This separation also permits the coding assistant and the simulation agent to
use different reasoning models. For example, Claude Code can remain the
user-facing orchestration environment while \gls{moosenger} performs
simulation-specific subtasks with a lower-cost hosted model or an
open-weight model. Such routing can reduce the amount of higher-cost
assistant-model inference devoted to domain-specific retrieval, input repair,
and repeated validation cycles while retaining the same harness-level
acceptance criteria. The current Claude Code plugin is a first step toward
this interaction model. Future work will extend the approach to additional
assistant and development environments and improve configuration, model
selection, and session-management interfaces, based on user feedback and
operational experience.

\section{Evaluation Methodology}
\label{sec:evaluation-methodology}

We evaluate \gls{moosenger} at progressively stronger levels of assurance. The 200-prompt benchmark separates the effects of the surrounding harness and the underlying reasoning model. Comparing each model with and without
\gls{moosenger} isolates the contribution of simulation-aware harnessing to
executable-input reliability, while comparing a hosted frontier model with a
locally deployable open-weight model under the same harness shows the remaining dependence on model capability. The complementary \gls{mms} benchmark adds a stronger assurance level by evaluating whether generated inputs remain
semantically aligned with the prescribed physics and reproduce known
analytical solutions when executability alone is insufficient. Together, these experiments distinguish the contributions stemming from system composition, model
capability, and the strength of the applied acceptance criterion.

\subsection{200-Prompt MOOSE Authoring Benchmark}
\label{sec:runnable-benchmark}

The benchmark contains eight MOOSE problem families, each with 25 natural-language
prompts, meaning 200 prompts in total. Each prompt requests a
complete runnable setup and specifies some combination of physics, geometry,
boundary or initial conditions, material parameters, and required outputs.
Table~\ref{tab:benchmark-families} summarizes the scope of each family. The
complete prompts are provided in Appendix~A.

\begin{table}[htbp]
\caption{Problem families in the 200-prompt MOOSE authoring benchmark. Each
family contains 25 prompts.}
\label{tab:benchmark-families}
\centering
\footnotesize
\renewcommand{\arraystretch}{1.16}
\setlength{\tabcolsep}{6pt}

\begin{tabularx}{\linewidth}{
    @{}
    >{\raggedright\arraybackslash}p{0.25\linewidth}
    >{\raggedright\arraybackslash}X
    @{}
}
\toprule
\textbf{Problem family} &
\textbf{Prompt coverage} \\
\midrule

\textbf{Diffusion} &
Steady and transient diffusion on 1D and 2D domains, with
standard boundary conditions and lightweight solution outputs. \\
\addlinespace[3pt]

\textbf{Transient heat conduction} &
Time-dependent conduction with explicit material properties, time-varying
boundary conditions, and transient execution requirements. \\
\addlinespace[3pt]

\textbf{Solid mechanics} &
Small-strain elasticity in one, two, and three dimensions, with displacement or
traction constraints and stress or reaction outputs. \\
\addlinespace[3pt]

\textbf{Porous flow} &
Darcy and pressure-diffusion problems with permeability, viscosity, porosity,
mixed boundary conditions, and selected verification quantities. \\
\addlinespace[3pt]

\textbf{Navier--Stokes} &
Steady and transient incompressible-flow problems requiring coupled velocity--pressure
formulations and appropriate solver configurations. \\
\addlinespace[3pt]

\textbf{Phase field} &
Allen--Cahn and Cahn--Hilliard interface-evolution or coarsening problems with
varied initial conditions, boundary conditions, and free-energy outputs. \\
\addlinespace[3pt]

\textbf{Plasticity} &
1D \(J_2\) elastoplastic bar problems spanning tension,
compression, load--unload, and traction-controlled cases, with stress--strain
and plastic-strain outputs. \\
\addlinespace[3pt]

\textbf{Reactor mesh generation} &
Reactor-oriented mesh-construction tasks involving geometry composition,
mesh-generation syntax, and boundary or subdomain definitions. \\

\bottomrule
\end{tabularx}
\end{table}

\subsection{Harness--Model Comparison and Executable Success Protocol}
\label{sec:agent-evaluation}
For each prompt, we evaluate four configurations under the same final execution gate: \gls{moosenger} with GPT~5.2 \gls{api}, \gls{moosenger} with Gemma~4~31B, standalone GPT~5.2 \gls{api} generation, and standalone Gemma 4 31B generation. Harness-enabled runs use the configured retrieval, MOOSE input sanitation/repair/validation, persistent workspace, and execution tools. Standalone runs receive the same natural-language prompt but do not call \gls{moosenger} tools during generation. The final candidate input is executed with \texttt{mcp.run\_input}; a prompt passes only when the authoritative runner reports \texttt{exit\_code=0}.

Each prompt is executed in an isolated workspace to avoid cross-case contamination and retain the final input, execution logs, and tool traces when enabled. This artifact-level record supports regression analysis and diagnosis of failures involving syntax, object types, parameters, symbol references, or runtime behavior. The four-way design separates two effects that would otherwise be conflated: \emph{same model, different system} measures the contribution of the harness, whereas \emph{same system, different model} measures residual dependence on reasoning-model capability.

\subsection{Methods of Manufactured Solutions Physics-Verification Benchmark}
\label{sec:mms-evaluation}

The \gls{mms} benchmark tests whether generated inputs go beyond executability and reproduce prescribed analytical solutions. The ten cases span steady and transient diffusion, heat conduction, Allen-Cahn phase-field evolution, Darcy flow, linear elasticity, incompressible Navier--Stokes flow, and coupled thermoelasticity. All cases use a 2D unit-square domain, $\Omega = [0,1]^2$.

Each case is constructed solution-first. A smooth analytical field
$\phi^{*}$ is prescribed, and the corresponding volumetric source or body-force terms are obtained by substituting the manufactured fields into the governing operators. These terms are derived symbolically with \texttt{SymPy} during benchmark generation and stored with each case. The actual evaluation prompts provide the resulting scalar or component-wise expressions in directly implementable form; the agent is therefore not required to perform the symbolic differentiation needed to construct the manufactured problem. This design isolates the skill being tested: namely, whether the agent can translate a fully specified mathematical problem into a correct, runnable MOOSE input file by selecting and configuring the appropriate module, variables, physics objects, materials, functions, boundary and initial conditions, and output postprocessors.

~\ref{app:mms-problems} presents the governing equations, parameters,
manufactured solutions, source definitions, and boundary and initial conditions for the ten cases. For readability, Table~\ref{tab:mms-cases} retains compact operator notation for the Allen-Cahn source term and for the elasticity, Navier--Stokes, and thermoelastic body-force terms. These compact expressions define the same manufactured forcing used by the benchmark but do not reproduce the fully expanded scalar or componentwise expressions supplied in the actual evaluation prompts. Each generated input is assessed using two complementary criteria: executable numerical accuracy against the manufactured solution and semantic alignment with the requested problem specification.

\subsubsection{Case Specification and Agent Prompt}
\label{sec:mms-prompt}

Each case is delivered to the agent as a single self-contained prompt that
states the full manufactured problem. The prompt provides, in structured form:
(i)~the governing equation in strong form; (ii)~the domain and dimension; 
(iii)~the primary solution variables; (iv)~the material constants; 
(v)~the prescribed analytical solution $\phi^{*}$ for each variable;
(vi)~the pre-computed source/forcing terms; (vii)~the boundary conditions and,
where required, additional constraints; (viii)~the initial conditions for
transient cases; and (ix)~the exact \texttt{[Postprocessors]} and
\texttt{[Outputs]} blocks to include verbatim (see
Listing~\ref{lst:mms-prompt-dirichlet}). The postprocessor block requests an
\texttt{ElementL2Error} for each primary variable, evaluated against the
manufactured solution and written to a CSV file (\texttt{mms\_results}) that the
numerical evaluator later reads. The prompt fixes neither the mesh resolution
nor the finite-element order; selecting a suitable discretization is part of the
agent's task, which is why the evaluation reports absolute error at a single
generated discretization rather than an asymptotic convergence rate
(cf.\ Section~\ref{sec:results}).

As a concrete example, the steady Dirichlet diffusion case supplies the strong
form $-\nabla\!\cdot\!(D\,\nabla u) = f$ on $\Omega=[0,1]^2$ with $D=1$, the
manufactured solution $u^{*}=\sin(\pi x)\sin(\pi y)$, the compatible source
$f = 2\pi^{2}\sin(\pi x)\sin(\pi y)$ obtained by substitution, and homogeneous
Dirichlet data $u=0$ on $\partial\Omega$. For the non-Dirichlet variants, the
prompt specifies the natural boundary treatment explicitly: the pure-Neumann
diffusion case gives $u^{*}=\cos(\pi x)\cos(\pi y)+2$ with source
$f = 2\pi^{2}\cos(\pi x)\cos(\pi y)$, which integrates to zero over $\Omega$
(satisfying the pure-Neumann compatibility condition), homogeneous natural
boundaries $\nabla u\cdot\mathbf{n}=0$ on all sides, and a domain-average
constraint $\langle u\rangle_{\Omega}=2$ that removes the constant nullspace and
selects the intended member of the solution family. Constructing an input that
correctly imposes such a constraint---rather than defaulting to a Dirichlet
condition---is exactly the kind of physics-setup competency the benchmark
targets.

The benchmark evaluates three levels of prompt detail: \emph{explicit},
\emph{module-aware}, and \emph{high-level}. These differ only in terms of amount of
introductory and module-specific guidance. All variants provide the same
manufactured solution, source terms, boundary and initial conditions, and
required output blocks. The results reported here use the \emph{explicit}
variant, which provides the full specification and directs the agent to use
the supplied source terms without re-deriving them.

\subsubsection{Numerical Accuracy Criterion}

Each generated input is executed independently using the target MOOSE
application. Successful execution alone is insufficient for a physics pass. The
input must produce the requested CSV output containing an \texttt{ElementL2\allowbreak{}Error}
postprocessor for every primary solution variable. For a variable $\phi$, the
absolute spatial $L^{2}$ error is the square root of the integrated squared
discrepancy:
\begin{equation}
    E_{L^2}(\phi)
    =
    \sqrt{\,
        \int_{\Omega}
        \bigl(\phi_h-\phi^{*}\bigr)^{2}
        \,d\Omega
    \,}
    \;=\;
    \left[
        \int_{\Omega}
        \bigl(\phi_h-\phi^{*}\bigr)^{2}
        \,d\Omega
    \right]^{1/2},
    \label{eq:mms-l2}
\end{equation}
where $\phi_h$ is the finite-element solution and $\phi^{*}$ is the manufactured
solution. This is exactly the quantity returned by the MOOSE
\texttt{ElementL2Error} postprocessor, which the evaluator reads directly from
the CSV file. For transient problems, the value is taken from the final nonempty
row of the CSV.

A case passes the physics evaluation only when the authoritative MOOSE execution
returns an exit code of zero and every required error satisfies:
\begin{equation}
    E_{L^2}(\phi)\leq \varepsilon_{\mathrm{tol}},
    \label{eq:mms-tolerance}
\end{equation}
with a shared tolerance $\varepsilon_{\mathrm{tol}}=0.1$. For multiphysics
problems, this criterion must hold independently for every requested variable. A
nonzero MOOSE exit code causes the case to fail before any CSV values are
accepted, even if a partial output file was written during the unsuccessful
solve.

\subsubsection{Semantic Alignment Criterion}

Physics execution and semantic alignment are evaluated separately. The alignment
evaluation compares the final generated MOOSE input with the original \gls{mms}
prompt and case specification. A language-model judge returns a structured
pass/fail decision along with an accompanying explanation. The primary criterion is
technical-scope alignment: namely, whether the generated input represents the requested
physical problem and its essential modeling structure. The judge assesses four
dimensions:

\begin{itemize}
\item Physical formulation: the governing equation and requested
physical model
\item Computational domain: the prescribed 2D
unit-square geometry
\item Model specification: the required variables, material
constants, source terms, boundary and initial conditions, and constraints
explicitly stated in the case
\item MOOSE realization: the intended MOOSE module and the
appropriate object families for representing the formulation.
\end{itemize}

Only requirements explicitly stated in the prompt or case specification are
treated as binding. Inferred or ambiguous details are treated as advisory. The
judge is also deliberately permissive about reporting and output configuration.
A missing, disabled, or partially incorrect postprocessor or output block does
not by itself cause an alignment failure when the requested physical formulation
and core model structure remain represented.

Semantic alignment therefore assesses whether an input expresses the requested
modeling intent, not whether it is structurally valid, executable, or numerically
accurate. These stronger claims are evaluated separately through authoritative
MOOSE execution and the numerical $L^2$-error criterion. An input may consequently
pass semantic alignment while failing a later gate because of an incorrectly
implemented constraint, an incompatible MOOSE parameter type, or solver
nonconvergence. The failures discussed in Section~\ref{sec:results} illustrate
this distinction.

\section{Results}
\label{sec:results}

\subsection{Simulation-Aware Harnessing Improves Executable Reliability}
The 200-prompt comparison in Table~\ref{tab:results_full_comparison} shows a large system-level effect. With GPT~5.2 \gls{api}, executable success increases from 10/200 (5\%) in standalone generation to 179/200 (89.5\%) with the \gls{moosenger} harness. With Gemma~4~31B, standalone generation succeeds on 0/200 cases, whereas the same model reaches 153/200 (76.5\%) when embedded in \gls{moosenger}. Thus, the benchmark indicates that model capability remains important, but the surrounding simulation-aware harness substantially alters the practical reliability of both model classes.

Across problem families, \gls{moosenger} + GPT~5.2 \gls{api} achieves 84\%--100\% executable success, including 100\% on diffusion and 96\% on solid mechanics. \gls{moosenger} + Gemma~4~31B reaches 60\%--92\%, with the lowest performance in regard to plasticity and Navier--Stokes. By contrast, standalone GPT~5.2 \gls{api} succeeds primarily on cases involving diffusion (36\%) and a single porous flow case (4\%), while standalone Gemma~4~31B does not produce an executable input in the benchmark. These family-level differences show that the harness benefit is not confined to a single physics class and that more structured multiphysics tasks remain more challenging even with tool support.

\begin{table*}[t]
\caption{End-to-end results for the 200-prompt \gls{moose} benchmark:
harness-enabled (\gls{moosenger}) vs.\ standalone LLM performance,
across GPT~5.2 \gls{api} and Gemma 4~31B backends.}
\label{tab:results_full_comparison}
\centering

\small
\renewcommand{\arraystretch}{1.10}
\setlength{\tabcolsep}{4pt}

\begin{adjustbox}{max width=\textwidth}
\begin{tabular}{@{} l c c c c @{}}
\toprule
\textbf{Problem family} &
\makecell{\textbf{\gls{moosenger}}\\\textbf{+ Gemma~4~31B}} &
\makecell{\textbf{\gls{moosenger}}\\\textbf{+ GPT 5.2}} &
\makecell{\textbf{ChatGPT 5.2 API}\\\textbf{standalone}} &
\makecell{\textbf{Gemma~4~31B}\\\textbf{standalone}} \\
\midrule

Diffusion &
\makecell[c]{23/25\\(92\%)} &
\makecell[c]{25/25\\(100\%)} &
\makecell[c]{9/25\\(36\%)} &
\makecell[c]{0/25\\(0\%)} \\

Transient heat conduction &
\makecell[c]{21/25\\(84\%)} &
\makecell[c]{23/25\\(92\%)} &
\makecell[c]{0/25\\(0\%)} &
\makecell[c]{0/25\\(0\%)} \\

Solid mechanics &
\makecell[c]{22/25\\(88\%)} &
\makecell[c]{24/25\\(96\%)} &
\makecell[c]{0/25\\(0\%)} &
\makecell[c]{0/25\\(0\%)} \\

Plasticity &
\makecell[c]{15/25\\(60\%)} &
\makecell[c]{21/25\\(84\%)} &
\makecell[c]{0/25\\(0\%)} &
\makecell[c]{0/25\\(0\%)} \\

Porous flow &
\makecell[c]{17/25\\(68\%)} &
\makecell[c]{23/25\\(92\%)} &
\makecell[c]{1/25\\(4\%)} &
\makecell[c]{0/25\\(0\%)} \\

Navier--Stokes &
\makecell[c]{16/25\\(64\%)} &
\makecell[c]{21/25\\(84\%)} &
\makecell[c]{0/25\\(0\%)} &
\makecell[c]{0/25\\(0\%)} \\

Phase field &
\makecell[c]{17/25\\(68\%)} &
\makecell[c]{21/25\\(84\%)} &
\makecell[c]{0/25\\(0\%)} &
\makecell[c]{0/25\\(0\%)} \\

Reactor mesh generation &
\makecell[c]{22/25\\(88\%)} &
\makecell[c]{21/25\\(84\%)} &
\makecell[c]{0/25\\(0\%)} &
\makecell[c]{0/25\\(0\%)} \\

\midrule
\textbf{Overall} &
\makecell[c]{\textbf{153/200}\\\textbf{(76.5\%)}} &
\makecell[c]{\textbf{179/200}\\\textbf{(89.5\%)}} &
\makecell[c]{\textbf{10/200}\\\textbf{(5\%)}} &
\makecell[c]{\textbf{0/200}\\\textbf{(0\%)}} \\
\bottomrule
\end{tabular}
\end{adjustbox}

\end{table*}

\subsection{Failure Recovery and Computational Trade-Offs}
The improvement is consistent with closing the loop between generation and executable evidence. Rather than requiring a correct one-shot draft, \gls{moosenger} can sanitize and normalize generated text, repair malformed \gls{hit} structure, replace invalid object types by using application syntax information, validate cross-block references and parameters, and return MOOSE diagnostics to the model for targeted revision. This converts executability from a one-shot generation property into an iterative artifact-repair process with observable intermediate signals.

The additional assurance also introduces computational overhead because harness-enabled runs perform retrieval, validation, repair, and, when appropriate, solver execution. The present benchmark is designed to measure reliability rather than optimize latency, so the principal comparison is pass rate under a common solver gate. Future evaluation will report tool-call counts, inference time, solver time, and recovery depth alongside success so reliability gains can be weighed against computational cost.

\subsection{Method of Manufactured Solutions Semantic Alignment and Numerical Accuracy}

Table~\ref{tab:mms_results} summarizes the ten benchmark cases, all generated by \gls{moosenger} using Gemma~4~31B as the underlying language model. The evaluation considers two criteria: semantic alignment with the requested problem specification and executable physics verification through MOOSE. All ten cases satisfied the semantic-alignment criterion, while eight also executed successfully and met the prescribed numerical-accuracy threshold. Because overall success requires both criteria, the two physics failures reduce the combined pass rate to \(80\%\).

The two physics failures illustrate why semantic alignment and executable verification are complementary to each other. The steady pure-Neumann diffusion input contained the requested diffusion equation, compatible source, homogeneous Neumann boundaries, and mean-value constraint, and therefore passed the alignment evaluation. Its authoritative MOOSE execution nevertheless terminated after 50 nonlinear iterations with \texttt{DIVERGED\_MAX\_IT}. Manual investigation confirmed that the mean-value constraint was not configured correctly. Because the process returned exit code~1, its partially generated CSV was not treated as a valid verification result.

The Navier--Stokes input similarly contained the requested velocity and pressure variables, forcing functions, boundary conditions, and error postprocessors, resulting in an alignment pass. However, the generated input supplied function names such as \texttt{u\_sol}, \texttt{v\_sol}, and \texttt{p\_sol} to numeric \texttt{DirichletBC\allowbreak{}/value} parameters. MOOSE consequently reported invalid syntax for these double-valued parameters and exited with code~1 before solving the problem. A function-capable boundary-condition object or the corresponding \texttt{function} parameter was required.

\begin{table}[htbp]
\caption{Results of the ten-case \gls{mms} evaluation performed by \gls{moosenger} using
\texttt{Gemma~4~31B}. Physics success requires successful MOOSE execution and
an accepted $L^2$ error below the prescribed threshold. A dash indicates that
no authoritative error was reported because MOOSE exited with a nonzero
status.}
\label{tab:mms_results}
\centering
\footnotesize
\renewcommand{\arraystretch}{1.12}
\setlength{\tabcolsep}{3pt}

\begin{tabularx}{\linewidth}{
    @{}
    >{\raggedright\arraybackslash}p{0.31\linewidth}
    >{\raggedright\arraybackslash}X
    >{\centering\arraybackslash}p{0.11\linewidth}
    >{\centering\arraybackslash}p{0.14\linewidth}
    @{}
}
\toprule
\textbf{Problem} &
\textbf{Final absolute $L^2$ error} &
\textbf{Physics} &
\textbf{Align.} \\
\midrule

Steady diffusion (Dirichlet) &
$E_{L^2}(u)=1.6448\times10^{-4}$ &
Pass & Pass \\

Steady diffusion (Neumann) &
\textemdash{} execution failed &
Fail & Pass \\

Transient diffusion (Dirichlet) &
$E_{L^2}(u)=2.3225\times10^{-2}$ &
Pass & Pass \\

Transient diffusion (Neumann) &
$E_{L^2}(u)=1.3523\times10^{-3}$ &
Pass & Pass \\

Allen--Cahn phase field &
$E_{L^2}(\eta)=1.6958\times10^{-2}$ &
Pass & Pass \\

Single-phase Darcy pressure &
$E_{L^2}(p)=1.6435\times10^{-4}$ &
Pass & Pass \\

Transient heat conduction &
$E_{L^2}(T)=1.1006\times10^{-4}$ &
Pass & Pass \\

Small-strain linear elasticity &
\makecell[l]{
    $E_{L^2}(d_x)=1.8020\times10^{-4}$\\
    $E_{L^2}(d_y)=2.6999\times10^{-4}$
} &
Pass & Pass \\

Steady incompressible Navier--Stokes &
\textemdash{} execution failed &
Fail & Pass \\

Steady thermoelasticity &
\makecell[l]{
    $E_{L^2}(T)=4.5686\times10^{-4}$\\
    $E_{L^2}(d_x)=8.6837\times10^{-4}$\\
    $E_{L^2}(d_y)=1.0326\times10^{-3}$
} &
Pass & Pass \\

\midrule
\textbf{Overall} &
\makecell[l]{
    \textbf{All accepted}\\
    \textbf{$L^2$ errors $<0.1$}
} &
\textbf{8/10} &
\textbf{10/10} \\
\bottomrule
\end{tabularx}
\end{table}

The successful cases produced errors of between approximately
\(1.1\times10^{-4}\) and \(2.3\times10^{-2}\), all comfortably below the specified tolerance. Despite relying on a comparatively limited non-frontier base model, \gls{moosenger} produced numerically consistent inputs for eight problems spanning several physical formulations, including nonlinear phase-field evolution and coupled thermoelasticity. This result provides evidence that the MOOSE-specific harness can compensate for some limitations of the underlying model by grounding its responses, exposing authoritative diagnostics, and supporting validation and repair.

The \(10/10\) semantic-alignment score indicates that the generated inputs represented the requested physical formulation and core model structure. However, the two execution failures show that semantic completeness alone does not guarantee a usable simulation. Taken together, these results demonstrate why \gls{moosenger} treats semantic alignment, executability, and numerical accuracy as distinct assurance gates. The \(8/10\) combined pass rate provides evidence that the framework can move beyond plausible input generation and toward executable, physics-informed verification, while the two failures expose remaining limitations that only authoritative solver-based checks can reveal.

This benchmark evaluates a single generated discretization for each problem against a fixed absolute-error threshold. It therefore provides an end-to-end assessment of solution accuracy but does not constitute a formal mesh-convergence study. Future work will extend the evaluation to sequences of uniformly refined meshes to verify both the expected asymptotic convergence rate and the prescribed absolute \(L^2\)-error tolerance.

\section{Simulation Modeling Workflow: Hexagonal Duct Bowing Benchmark}
\label{sec:vtb-demo}

To illustrate the architecture on a realistic engineering workflow, we use a sodium fast reactor hexagonal duct bowing model from the \gls{vtb}. This case is an International Atomic Energy Agency benchmark that examines free thermal bowing of a mechanically fixed 3D hexagonal duct subjected to radial and axial thermal
gradients. The published case uses the MOOSE Reactor-module mesh generation and the Solid Mechanics module to calculate thermomechanical deformation~\cite{moose_tm_assess_2021}. This case is well suited for a workflow demonstration because it exercises MOOSE input features that are challenging for an agent, including coupled thermal and mechanical physics, functions and auxiliary variables, Dirichlet boundary conditions, SolidMechanics physics syntax, material models, and the postprocessing needed to recover the benchmark response. Gemma~4~31B is used as the underlying language model in \gls{moosenger}.

\begin{figure*}[!tp]
  \centering
  \includegraphics[width=0.94\textwidth]{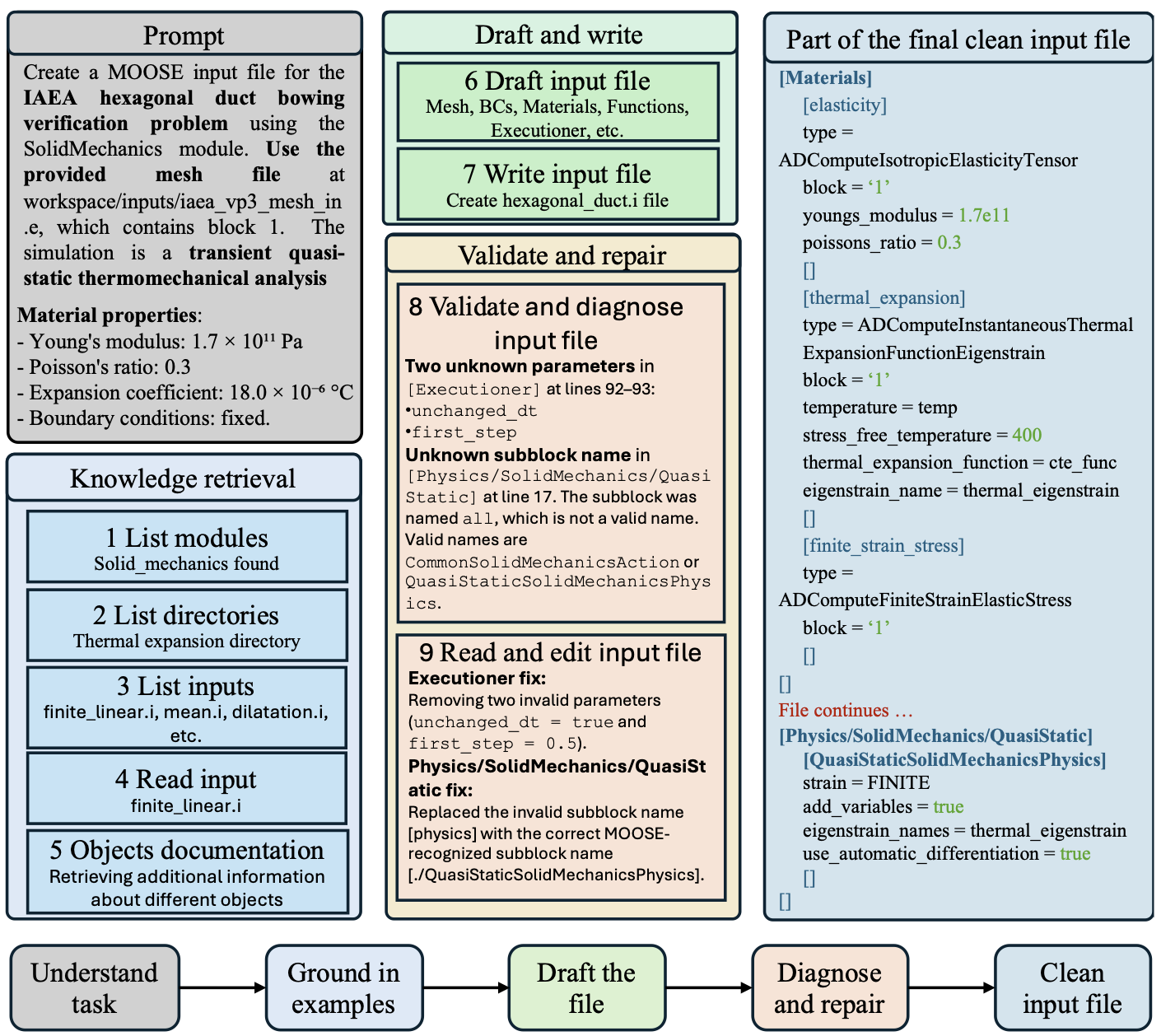}
  \caption[MOOSEnger evidence trail for the VTB hexagonal duct-bowing case.]{%
  \gls{moosenger} evidence trail for the \gls{vtb} hexagonal duct-bowing
  case. The figure shows the original problem specification, knowledge
  grounding, input drafting, validation diagnostics, repair, and the resulting
  clean MOOSE input artifact. The final simulation comparison is shown
  separately in Figure~\ref{fig:duct_bowing_results} so that the agent workflow
  and numerical outcome can be inspected independently.}
  \label{fig:duct_bowing_workflow}
\end{figure*}

Figure~\ref{fig:duct_bowing_workflow} shows the agent workflow. The case begins from a natural-language request that specifies the benchmark problem, provided mesh, material properties, and fixed boundary conditions. \gls{moosenger} first interprets the task and grounds the model by listing relevant MOOSE modules, browsing thermal-expansion examples, reading candidate inputs, and retrieving object documentation. It then drafts and writes a persistent MOOSE input artifact before entering validation and
repair. The validator identifies unsupported \texttt{Executioner} parameters and an invalid subblock under \texttt{Physics/SolidMechanics/QuasiStatic}; the agent edits the same artifact and repeats validation until a clean input is obtained. 
This case is intended primarily as a workflow demonstration rather than a test of fully independent model construction, since the agent is allowed to inspect a published reference input during grounding.

\begin{figure*}[!tp]
  \centering

  \begin{subfigure}[t]{0.42\textwidth}
    \centering
    \includegraphics[height=0.40\textheight]
      {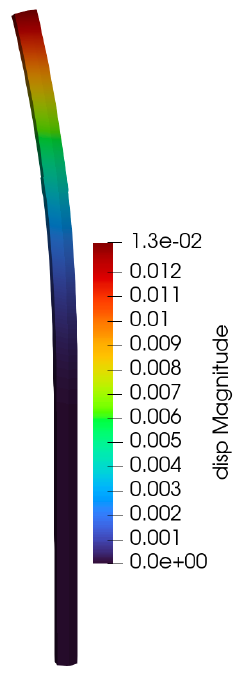}
    \caption{MOOSEnger-generated model}
    \label{fig:duct-bowing-result-moosenger}
  \end{subfigure}
  \hspace{0.06\textwidth}
  \begin{subfigure}[t]{0.42\textwidth}
    \centering
    \includegraphics[height=0.40\textheight]
      {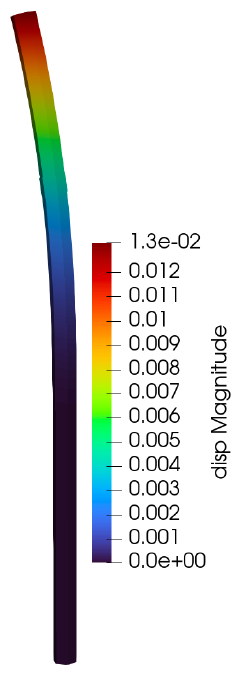}
    \caption{Reference model}
    \label{fig:duct-bowing-result-reference}
  \end{subfigure}

  \vspace{0.5em}

  \begin{minipage}{0.52\textwidth}
    \centering
    \small
    \begin{tabular}{@{}l r@{}}
      \toprule
      \textbf{Workflow metric} & \textbf{Value} \\
      \midrule
      Tool calls        & 25 \\
      Repair iterations & 3 \\
      Errors resolved   & 3 \\
      Duration          & 85.6\,s \\
      \bottomrule
    \end{tabular}
  \end{minipage}

  \caption{Final result comparison for the International Atomic Energy Agency hexagonal duct-bowing
  demonstration. (a) Result obtained from the MOOSEnger-generated and
  repaired input and (b) the published reference model. The workflow summary
  reports the agent effort required to obtain the accepted simulation input.}
  \label{fig:duct_bowing_results}
\end{figure*}

Across the complete workflow, \gls{moosenger} made 25 tool calls, 
required three validation-and-repair iterations, resolved three detected input errors, and completed the case in 85.6~s. The successful completion of this nontrivial workflow with an open-weight, non-frontier model provides strong case-level evidence
of the \gls{moosenger} harness's effectiveness: reliability is supported not only by the underlying language model, but also by knowledge grounding, persistent artifact management, MOOSE-specific validation, and iterative repair. Figure~\ref{fig:duct_bowing_results} presents the final simulation output separately from the agent workflow. The \gls{moosenger}-generated model and the reference model exhibit the same characteristic bowed duct shape and are plotted using the same displacement-magnitude scale. The comparison thus provides evidence that the harness-guided workflow can recover from detected input errors and produce a simulation artifact consistent with the expected benchmark deformation pattern.

\section{Discussion}
\label{sec:discussion}

\subsection{Why Simulation-Aware Harnessing Changes Reliability}
The four-way benchmark supports a systems interpretation of agent reliability.
When holding the model family fixed, adding the simulation-aware harness greatly
improves the executable success for both evaluated models. The remaining
difference between the harness-enabled models shows that the surrounding system
does not eliminate model-capability differences; instead, it constrains and
amplifies model capability by supplying domain evidence, persistent artifacts,
deterministic checks, and executable feedback.

Several implementation lessons follow from this result. First, domain grounding must provide authoritative construction knowledge rather than merely topically relevant text. \gls{moosenger} preserves the block hierarchy and object--parameter relationships found in documentation and examples, and augments them with MOOSE-native metadata, including registered object types, parameter schemas, data types, default values, and required-parameter status. This structured evidence helps the model select and configure appropriate objects during input authoring. Second, the resulting artifact must be evaluated through deterministic mechanisms independent of the model's own assessment. Formal \gls{hit} parsing, registry-conformance and reference checks, language-server diagnostics, and executable MOOSE validation provide concrete evidence that the model can interpret when proposing repairs. Third, stable subtasks benefit from explicit workflow control. \gls{moosenger} governs pre-execution validation and repair through a bounded, stateful graph that consistently applies the relevant deterministic checks and routes failures through controlled recovery, yet retains agent autonomy for retrieval, model construction, planning, and interpretation of diagnostic evidence.

\subsection{Extending Artifact Assurance Toward Engineering Verification and Validation}

The assurance ladder distinguishes the progressively stronger claims supported by different forms of simulation evidence. Structural checks establish the internal consistency of an input, executable checks determine whether the target MOOSE application can process and run it, and semantic and numerical checks provide stronger evidence about the encoded formulation and computed solution. Passing one gate supports only the claim associated with that gate and does not imply satisfaction of subsequent levels. \gls{moosenger} operationalizes this distinction through configured acceptance gates applied to explicitly identified artifact revisions.

Extending this architecture toward engineering \gls{vv} is a natural progression of \gls{moosenger}'s existing artifact-governed validation-and-repair process. The configured acceptance gates could be expanded with application-specific criteria covering geometry, material properties, constitutive models, boundary and initial conditions, multiphysics couplings, and quantities of interest. Before execution, these criteria could be used to assess whether the input artifact faithfully represents the intended engineering problem and trace each important requirement to its implementation in the model. After execution, solution-verification criteria could assess mesh- and time-step-refinement behavior, conservation and balance errors, solver convergence, numerical uncertainty, and agreement with analytical solutions or established numerical benchmarks. Validation criteria could further evaluate selected quantities of interest against experimental data or other qualified reference evidence, examine expected physical trends, account for relevant input uncertainties, and assess whether the requested prediction lies within the validated application domain.

These additional assessments would fall under the same artifact-governed lifecycle as the current structural and executable checks. Each result would be bound to the exact candidate revision that was evaluated. A failed criterion could return structured evidence to the agent for targeted repair, additional evidence gathering, or controlled replanning, after which the revised artifact would be independently reevaluated. Repeated or unresolved failures could
trigger bounded termination or escalation for expert review. A successful evaluation could retain the accepted input, solver configuration, reference evidence, assessment results, and acceptance thresholds as a traceable evidence package. In this way, physics-specific assessment would become part of the iterative acceptance process rather than remaining a separate evaluation performed only after the agent workflow is complete.

This extension would not replace engineering judgment or independently establish regulatory-grade \gls{vv}. Qualified users would remain responsible for selecting appropriate models, reference evidence, uncertainty treatments, and acceptance criteria. The role of \gls{moosenger} would be to make those criteria explicit, apply them consistently via its existing artifact-governed workflow, preserve their provenance, and expose the resulting evidence for expert review.

\subsection{Model Portability and Deployment Trade-Offs}

\gls{moosenger}'s model-agnostic design makes the surrounding knowledge,
validation, artifact-governance, and execution contracts reusable across
hosted and locally deployable models. The Gemma~4 31B result is important in
this context: although it remains below GPT~5.2 \gls{api} under the same harness,
76.5\% executable success is a substantial change from its 0\% standalone
result. This suggests a practical deployment pattern in which organizations
can select reasoning models according to capability, cost, data-control, or
local-compute requirements while still retaining the same simulation-aware
validation and execution layer. Model substitution should nevertheless be
evaluated under common artifact-level gates because prompt quality alone does
not reveal whether a replacement preserves scientific workflow reliability.

Interface portability extends the same model-agnostic design. As described in
Section~\ref{sec:interfaces}, \gls{moosenger} can support different user-facing
environments while preserving the same simulation-aware grounding, validation,
artifact governance, and execution layer. This separation also allows the
interface and simulation agent to use different models, including lower-cost
hosted or open-weight models for domain-specific tasks. Together, model and
interface portability reduce dependence on any particular language model or
interaction environment.

\section{Conclusion and Future Work}
\label{sec:conclusion}

We presented \gls{moosenger}, an modeling and simulation AI agent framework
built around a simulation-aware harness for the MOOSE ecosystem. Rather than
treating simulation input authoring as one-shot text generation, \gls{moosenger}
manages it as a revisioned scientific artifact whose acceptance depends
on explicit evidence. The reasoning model interprets user intent and proposes
changes, while the harness provides MOOSE-specific grounding, maintains
artifact identity, binds validation and execution evidence to exact revisions,
controls repair and promotion, and obtains authoritative feedback from the
target MOOSE application. This separation allows reasoning models, execution
backends, and user-facing interfaces to change without altering the
artifact-level conditions used to accept a simulation input.

The evaluation demonstrates the resulting system-level improvement at
progressively stronger levels of assurance. On the 200-prompt benchmark,
\gls{moosenger} increased executable success from \(5\%\) to \(89.5\%\) for
GPT~5.2 \gls{api} and from \(0\%\) to \(76.5\%\) for Gemma~4~31B. The difference
between the two harness-enabled models shows that model capability remains
important, while the large gains over standalone generation demonstrate the
effect of the surrounding simulation-aware system. The complementary
\gls{mms} benchmark extended the evaluation beyond executability: all ten
cases passed semantic alignment, and eight also executed successfully while
meeting the prescribed single-mesh \(L^2\)-error criterion. The
reference-grounded \gls{vtb} duct-bowing case further demonstrated the same
artifact-governed workflow in a coupled engineering problem, including
iterative recovery from multiple input errors and a final deformation pattern
consistent with the published reference result. Together, these experiments
support the central conclusion that reliable simulation assistance depends on
the complete agent system rather than on the language model alone.

These results establish an initial assurance foundation rather than complete
engineering \gls{vv}. The current gates provide evidence for input consistency,
target-application executability, semantic alignment, and, for the
\gls{mms} cases, numerical agreement with prescribed analytical solutions.
Because the same artifact-governed lifecycle can incorporate additional
revision-bound evidence, \gls{moosenger} provides a framework for extending
these checks toward application-specific \gls{vv} while still
retaining qualified users as the authority for modeling assumptions,
acceptance criteria, interpretation of evidence, and final engineering
decisions.

A primary direction for future work is therefore to strengthen these
application--specific assurance gates through concrete verification studies.
Planned work includes mesh-refinement studies for the \gls{mms} benchmark,
requirement traceability for representative engineering models, and
post-execution assessments of conservation, mesh- and time-step sensitivity,
numerical uncertainty, and agreement with qualified reference evidence.
These assessments can use the existing revision-bound workflow to trigger
additional evidence gathering, targeted repair, controlled replanning, or
escalation when their criteria are not satisfied.

Future evaluation will also broaden application coverage and more precisely
characterize cases in which reliability gains arise. Controlled ablations and stronger
baselines will isolate the contributions of retrieval, deterministic syntax
and reference checks, language-server diagnostics, iterative repair, and
executable validation. Benchmark reporting will expand beyond executable
success to include semantic and numerical criteria, failure stage and root
cause, repair depth, tool calls, inference and solver time, and computational
cost, together with clearer documentation of prompt construction, retrieved
example overlap, and software and model configurations needed for
reproducibility.

Finally, future work will connect persistent cross-run memory with the
existing governed knowledge-promotion pipeline. Promotion mechanisms for
structured knowledge candidates are already implemented, but information
from independent sessions is not yet routinely accumulated and converted
into such candidates. Integrating these mechanisms would allow recurring
diagnostics, verified repairs, and stable project knowledge to become
retrieval-visible only after conducting the appropriate provenance, freshness, and evidence
checks, while stale, failed, or superseded information remains excluded from
the active knowledge base.

\section{Acknowledgment}
This research was funded by the U.S. Department of Energy Office of Nuclear Energy (DOE-NE) NEAMS project under contract no.~DE-NE0008983. This research made use of the Idaho National Laboratory (INL) HPC systems located at the Collaborative Computing Center and supported by DOE-NE and the Nuclear Science User Facilities under contract no.~DE-AC07-05ID14517.

\section*{Code availability}
The \gls{moosenger} codebase is hosted on INL's GitLab and is currently undergoing laboratory open-source review. Access can be provided upon request during the review period.

\section*{Declaration of generative AI and AI-assisted technologies in the manuscript preparation process}
During preparation of this work, the authors used OpenAI ChatGPT 5.6 to improve the writing and formatting of the manuscript, and OpenAI Codex to assist with coding. After using this tool/service, the authors reviewed and edited the content as needed and take full responsibility for the content of the published article.

\label{sec:Reference}
\bibliographystyle{elsarticle-num-names}
\bibliography{main}

\appendix
\makeatletter
\renewcommand{\@seccntformat}[1]{\textbf{\csname the#1\endcsname}\quad}
\makeatother

\section{Agent Evaluation Prompts}
\label{app:agent-eval-prompts}
\noindent This appendix reproduces the 200 natural-language prompts used in
the agent evaluation. The benchmark contains eight problem families, with 25
prompts per family. Prompt wording is preserved from the evaluation JSON files;
mathematical symbols and special characters are escaped only for LaTeX
typesetting.

\subsection{Diffusion}
{\small\sloppy\setlength{\emergencystretch}{3em}%
\begin{enumerate}[label=D\arabic*., leftmargin=*, align=left, itemsep=0.3em, topsep=0.3em]
  \item Create a minimal 2D transient diffusion input with constant material properties. Use Dirichlet
    BC on the left (u=1) and right (u=0). Then run it.
  \item Write a simple 2D diffusion problem on a unit square with a constant source term and steady
    solve. Provide the full input and run it.
  \item Generate a 1D transient diffusion example on [0,1] with initial condition u=0, left boundary
    u=1, right boundary u=0. Run it.
  \item Make a 2D diffusion simulation using a generated mesh (e.g., GeneratedMesh). Use a linear solver
    and run the case.
  \item Produce a very small diffusion input that runs fast (coarse mesh, short end time). Include BCs
    and run it.
  \item Create a 2D diffusion on a rectangular mesh (nx=ny \ensuremath{\sim} 10). Use steady-state and
    Dirichlet BCs on all sides. Run it.
  \item Write a 2D diffusion problem with a Gaussian initial condition and transient time stepping. Run
    it.
  \item Generate an input for heat conduction (diffusion) in 2D with constant kappa. Use implicit Euler
    and run it.
  \item Make a diffusion example with a Neumann flux on the top boundary and Dirichlet u=0 elsewhere.
    Run it.
  \item Create a diffusion input that uses a Material to provide diffusivity. Run it.
  \item Write a 1D steady diffusion problem with u(0)=0 and u(1)=1. Use a minimal mesh and run it.
  \item Create a 2D transient diffusion input with u=0 initial condition and a constant volumetric
    source. Run it.
  \item Produce a tiny diffusion test case that outputs the solution field to Exodus. Run it.
  \item Generate a diffusion input that uses linear Lagrange FE, solve steady, and run.
  \item Make a 2D diffusion example on a 1x1 square using a coarse mesh, solve steady, and run it.
  \item Create a diffusion case with a time-dependent Dirichlet BC (e.g., u=sin(t)) on the left
    boundary; run a short transient.
  \item Write a transient diffusion problem with a small dt and short end time. Keep it minimal and run
    it.
  \item Generate a 2D diffusion input with separate blocks for Mesh, Variables, Kernels, BCs,
    Executioner, and Outputs. Run it.
  \item Create a simple diffusion example that uses PETSc linear solver defaults and runs.
  \item Make a diffusion case with adaptive time stepping if available, but keep it simple and runnable.
    Run it.
  \item Write a steady diffusion input with a source term and Dirichlet BCs; ensure it is runnable and then
    run it.
  \item Create a 2D diffusion input with a small mesh and transient solve; run it.
  \item Generate a 1D diffusion input that runs quickly and writes CSV output; run it.
  \item Make a diffusion example with u fixed on left/right and insulated top/bottom; run it.
  \item Create a minimal diffusion input that you expect to work with common MOOSE apps; run it.
\end{enumerate}}

\subsection{Transient heat conduction}
{\small\sloppy\setlength{\emergencystretch}{3em}%
\begin{enumerate}[label=H\arabic*., leftmargin=*, align=left, itemsep=0.3em, topsep=0.3em]
  \item Create a transient 1D heat conduction MOOSE case for a 0.5 m steel rod (k=45
    W/m\ensuremath{\cdot}K, \ensuremath{\rho}=7800 kg/m\textsuperscript{3}, cp=470
    J/kg\ensuremath{\cdot}K). Initial T=293 K. At t=0, set T=373 K at x=0 and T=293 K at x=L. Simulate
    0--200 s and output T(x,t) plus T at x=0.25 m vs time.
  \item Transient 1D aluminum rod L=0.2 m (k=205, \ensuremath{\rho}=2700, cp=900). Initial T=300 K.
    Apply heat flux q''=1e5 W/m\textsuperscript{2} at x=0 for 0<t<0.5 s then q''=0; set x=L insulated.
    Run 0--5 s and write temperature at x=0 and x=0.1 m to CSV.
  \item Model a 1D polymer slab L=0.1 m (k=0.2, \ensuremath{\rho}=1200, cp=1500). Initial T=373 K. Both
    ends convect to ambient 293 K with h=25 W/m\textsuperscript{2}\ensuremath{\cdot}K. Run 0--600 s and
    output average temperature vs time.
  \item 2D square plate 1\ensuremath{\times}1 m with k=10, \ensuremath{\rho}=5000, cp=400. Initial 293
    K. Left edge fixed at 400 K; other three edges convection to 293 K with h=15. Run 0--50 s (adaptive
    dt OK). Output Exodus and center temperature vs time.
  \item 2D rectangle 0.3\ensuremath{\times}0.1 m stainless steel (k=16, \ensuremath{\rho}=8000, cp=500).
    Initial 293 K. Add volumetric heat generation Q=2e6 W/m\textsuperscript{3} only in a centered
    0.05\ensuremath{\times}0.05 m region; outer boundaries insulated. Run 0--10 s and output max
    temperature vs time.
  \item 2D 0.2\ensuremath{\times}0.2 m plate split at x=0.1 m: left is copper (k=400,
    \ensuremath{\rho}=8960, cp=385), right is glass (k=1.2, \ensuremath{\rho}=2500, cp=800). Initial 293
    K. Left boundary T=373 K; right boundary insulated; top/bottom convection to 293 K with h=10. Run
    0--100 s and output temperature along the material interface.
  \item 2D 0.1\ensuremath{\times}0.1 m epoxy (k=0.3, \ensuremath{\rho}=1200, cp=1000) with a copper
    circular inclusion radius 0.015 m at the center. Initial 293 K. Deposit total power 10 W uniformly
    in the inclusion (convert to volumetric). All outer edges convect to 293 K with h=30. Run 0--30 s
    and output center temperature.
  \item 3D cube side 0.05 m, material k=150, \ensuremath{\rho}=2330, cp=700. Initial 300 K. Bottom face
    held at 350 K; all other faces convection to 300 K with h=100. Run 0--2 s and output average T and
    bottom-face heat flux vs time.
  \item 3D brick 0.1\ensuremath{\times}0.05\ensuremath{\times}0.02 m aluminum (k=205,
    \ensuremath{\rho}=2700, cp=900). Initial 300 K. Apply a spatial Gaussian volumetric heat source
    centered at (0.05,0.025,0.01) with total power 50 W and time decay exp(-t/0.2 s). All boundaries
    insulated. Run 0--1 s and output max temperature vs time.
  \item Axisymmetric cylinder: radius 0.01 m, height 0.05 m, k=20, \ensuremath{\rho}=7800, cp=500.
    Initial 293 K. Outer radius insulated, inner radius convection to 293 K with h=5, top face heat flux
    q''=2e4 W/m\textsuperscript{2} for 1 s then off. Run 0--10 s; output T at r=0, z=0.025.
  \item 2D 1\ensuremath{\times}1 m domain, anisotropic kx=50, ky=5 W/m\ensuremath{\cdot}K, with
    \ensuremath{\rho}=3000, cp=600. Initial 293 K. Left edge 350 K, right edge 293 K, top/bottom
    insulated. Run 0--20 s and output temperature contours (Exodus).
  \item 1D rod L=0.1 m with \ensuremath{\rho}=5000, cp=500 and k(T)=10*(1+0.01*(T-300)). Initial 300 K.
    Left boundary ramps 300\ensuremath{\rightarrow}600 K over 2 s; right boundary fixed 300 K. Run 0--5
    s and output profile snapshots.
  \item 2D 0.2\ensuremath{\times}0.2 m, k=15, \ensuremath{\rho}=7800, cp(T)=400+0.5*(T-300). Initial 300
    K. Convection on all sides to 300 K with h=50. Add uniform volumetric source Q=1e6
    W/m\textsuperscript{3} for 0<t<0.5 s then off. Run 0--5 s; output average temperature.
  \item 1D slab L=0.5 m with \ensuremath{\rho}=2000, cp=1000, and k(x)=1+9*(x/L) W/m\ensuremath{\cdot}K.
    Initial 300 K. Set x=0 to 400 K; set x=L insulated. Run 0--100 s; output T(x,t).
  \item 2D 0.1\ensuremath{\times}0.1 m, k=30, \ensuremath{\rho}=7800, cp=500. Initial 293 K. Left edge
    T=293+50*sin(2\ensuremath{\pi} t) K; other edges convection to 293 K with h=20. Run 0--5 s; output
    center temperature vs time.
  \item 1D rod L=0.2 m (k=100, \ensuremath{\rho}=8000, cp=400). Initial 293 K. At x=0 apply square-wave
    flux: q''=5e4 W/m\textsuperscript{2} for 0.1 s on / 0.1 s off repeating. At x=L fix T=293 K. Run
    0--2 s and output T at x=0.02 m.
  \item 2D 0.5\ensuremath{\times}0.5 m, k=5, \ensuremath{\rho}=2500, cp=800. Initial 350 K. All edges
    convect with h=10. Ambient temperature steps from 300 to 280 K at t=10 s. Run 0--60 s and output
    average temperature and total convective heat loss vs time.
  \item 1D slab L=0.05 m, k=2, \ensuremath{\rho}=1000, cp=2000. Initial 293 K. Left boundary held at 500
    K for t<1 s then instantly set to 293 K; right boundary insulated. Run 0--10 s; output mid-plane
    temperature vs time.
  \item 2D 0.2\ensuremath{\times}0.1 m, k=12, \ensuremath{\rho}=6000, cp=450. Initial 293 K. Left edge
    fixed 350 K; right edge insulated; bottom edge convection to 293 K with h=25; top edge prescribed
    heat flux q''=1e4 W/m\textsuperscript{2}. Run 0--30 s; output max T and boundary heat rates.
  \item 2D strip 1\ensuremath{\times}0.1 m, k=20, \ensuremath{\rho}=7800, cp=500. Initial 293 K. Left
    boundary 400 K, right boundary 293 K, top/bottom insulated. Use a non-uniform mesh refined near x=0
    to capture steep gradients. Run 0--5 s; output centerline T(x) at multiple times.
  \item 2D 0.2\ensuremath{\times}0.2 m, k=15, \ensuremath{\rho}=2700, cp=900. Initial condition:
    T=293+200*exp(-((x-0.1)\textsuperscript{2}+(y-0.1)\textsuperscript{2})/(2*0.01\textsuperscript{2})).
    All boundaries insulated. Run 0--1 s and output peak temperature decay vs time.
  \item 3D cube 0.1 m, k=40, \ensuremath{\rho}=7800, cp=500. Initial 293 K. Volumetric heat generation
    Q=0 for t<5 s, then Q=5e5 W/m\textsuperscript{3} for t\ensuremath{\ge}5 s. All faces convection to
    293 K with h=15. Run 0--30 s; output average and max T vs time.
  \item 2D 0.3\ensuremath{\times}0.3 m, k=8, \ensuremath{\rho}=3000, cp=700. Initial 293 K. Left edge
    323 K, right edge 293 K, top/bottom convection to 293 K with h=5. Add probe outputs for T at
    (0.15,0.15) and (0.25,0.15) to CSV. Run 0--200 s.
  \item 2D domain 0.2 m in x and 0.05 m in y, k=10, \ensuremath{\rho}=2500, cp=900. Make y boundaries
    periodic. At x=0 set T=350 K; at x=0.2 apply convection to 300 K with h=20. Initial 300 K. Run 0--50
    s and output y-averaged temperature along x (or sample along y=midline).
  \item 2D unit square with exact solution
    T(x,y,t)=300+10*sin(\ensuremath{\pi}x)*sin(\ensuremath{\pi}y)*exp(-t). Use k=1, \ensuremath{\rho}=1,
    cp=1 and add the needed volumetric source so this is an exact transient conduction solution. Apply
    Dirichlet BCs from the exact function. Run to t=1 and compute L2 error vs exact solution over
    time.
\end{enumerate}}

\subsection{Solid mechanics}
{\small\sloppy\setlength{\emergencystretch}{3em}%
\begin{enumerate}[label=S\arabic*., leftmargin=*, align=left, itemsep=0.3em, topsep=0.3em]
  \item Build and run a small-strain linear elastic 1D bar (L=1 m) with E=210 GPa, \ensuremath{\nu}=0.3.
    Fix x=0, prescribe u=1e-3 m at x=L. Output axial stress and reaction force vs load step.
  \item Build and run a small-strain linear elastic 1D bar (L=1 m, A=1 m\textsuperscript{2}) with E=100
    GPa, \ensuremath{\nu}=0.25. Fix x=0, apply a constant traction on x=L of 10 MPa. Output displacement
    at x=L and verify \ensuremath{\sigma}xx is \ensuremath{\sim}10 MPa everywhere.
  \item Create and run a 1D bar in compression (L=0.5 m) with E=70 GPa, \ensuremath{\nu}=0.33. Fix x=0,
    prescribe u=-5e-4 m at x=L. Output \ensuremath{\sigma}xx and reaction force at x=0.
  \item Create and run a 2D plane-stress rectangle (1.0\ensuremath{\times}0.2 m), linear elastic with
    E=70 GPa, \ensuremath{\nu}=0.33. Fix left edge (ux=uy=0). Prescribe uniform ux=1e-3 m on the right
    edge (leave uy free). Output displacement at (1,0.1) and \ensuremath{\sigma}xx field.
  \item Create and run a 2D plane-strain block (0.1\ensuremath{\times}0.1 m), linear elastic with E=10
    GPa, \ensuremath{\nu}=0.25. Bottom edge fixed (ux=uy=0). Prescribe uniform uy=-1e-4 m on the top
    edge (ux free). Sides traction-free. Output average \ensuremath{\sigma}yy and reaction force on the
    top edge.
  \item Build and run a 2D simple shear test on a 1\ensuremath{\times}1 m square (plane strain). Use E=1
    GPa, \ensuremath{\nu}=0.3. Fix bottom edge (ux=uy=0). Prescribe ux=0.01 m on the top edge and set
    uy=0 on the top edge. Output \ensuremath{\tau}xy at the center vs time/step.
  \item Create and run a 2D cantilever (L=2 m, H=0.2 m), plane stress, E=200 GPa, \ensuremath{\nu}=0.3.
    Clamp left edge (ux=uy=0). Apply a uniform downward traction on the right edge of 1 MPa. Output tip
    (right edge mid-height) vertical displacement and von Mises stress.
  \item Create and run a 2D cantilever (L=1 m, H=0.1 m), plane stress, E=100 GPa, \ensuremath{\nu}=0.3.
    Clamp left edge. Prescribe a uniform downward displacement uy=-1e-4 m on the right edge (ux free).
    Output reaction force on the right edge and max von Mises stress.
  \item Build and run a 2D square (1\ensuremath{\times}1 m), plane stress, E=10 GPa,
    \ensuremath{\nu}=0.3. Fix the bottom edge (uy=0) and pin one point on the bottom-left corner in x to
    remove rigid motion. Apply a uniform pressure (traction) of 0.5 MPa downward on the top edge. Output
    top-edge average uy and total reaction on the bottom.
  \item Create and run a 2D rectangle (1\ensuremath{\times}0.5 m), plane stress, E=50 GPa,
    \ensuremath{\nu}=0.25. Apply a uniform body force in -y direction (e.g., gravity-like) and add
    minimal constraints to prevent rigid motion (pin one node in x and one node in y). Output the
    displacement field and total reactions.
  \item Make a very simple 2D patch test for linear elasticity (plane stress): impose an exact linear
    displacement field (e.g., ux=a*x, uy=b*y) using Dirichlet BCs on all boundaries so strain is
    constant. Verify stress is uniform by reporting min/max \ensuremath{\sigma}xx and
    \ensuremath{\sigma}yy.
  \item Create and run a 2D plane-stress rectangle (1\ensuremath{\times}1 m), E=100 GPa,
    \ensuremath{\nu}=0.3. Apply ux=0 at x=0 and ux=1e-3 at x=1. Leave y-edges traction-free and
    constrain one point in y to remove rigid motion. Output average \ensuremath{\sigma}xx and lateral
    contraction (average uy on the top edge).
  \item Build and run a 2D two-material strip tension test (2\ensuremath{\times}0.2 m), plane stress.
    Left half: E=200 GPa, \ensuremath{\nu}=0.3. Right half: E=20 GPa, \ensuremath{\nu}=0.3. Fix left
    edge, prescribe ux=1e-3 m on right edge. Output \ensuremath{\sigma}xx along the centerline and the
    reaction force on the right edge.
  \item Build and run a 2D two-material strip compression test (1\ensuremath{\times}0.2 m), plane
    strain. Bottom half: E=100 GPa, \ensuremath{\nu}=0.25. Top half: E=10 GPa, \ensuremath{\nu}=0.25.
    Fix bottom edge, prescribe uy=-1e-4 m on top edge. Output \ensuremath{\sigma}yy field and compare
    average \ensuremath{\sigma}yy in each layer.
  \item Create and run a 2D plane-stress rectangle (1\ensuremath{\times}0.2 m), E=70 GPa,
    \ensuremath{\nu}=0.33. Fix left edge. Prescribe ux=1e-3 m on right edge. Postprocess an estimate of
    Poisson's ratio by measuring average \ensuremath{\varepsilon}xx and \ensuremath{\varepsilon}yy from
    displacements (report both).
  \item Create and run a 2D quarter-symmetry tension model: full plate would be 1\ensuremath{\times}1 m,
    but model only the top-right quarter [0.5,1]\ensuremath{\times}[0.5,1]. Apply symmetry BCs on x=0.5
    and y=0.5. Pull on x=1 with prescribed ux=1e-3 m. Use E=100 GPa, \ensuremath{\nu}=0.3 plane stress.
    Output displacement at (1,1) and \ensuremath{\sigma}xx field.
  \item Create and run an axisymmetric thick cylinder (2D r--z rectangle), small-strain linear elastic
    with E=210 GPa, \ensuremath{\nu}=0.3. Inner radius 0.05 m, outer radius 0.1 m, height 0.1 m. Apply
    internal pressure 10 MPa on r=0.05, outer surface traction-free. Add minimal z constraints to remove
    rigid motion. Output radial displacement at the inner wall and hoop stress
    \ensuremath{\sigma}\ensuremath{\theta}\ensuremath{\theta}.
  \item Create and run an axisymmetric solid rod in tension: radius 0.01 m, length 0.1 m, E=200 GPa,
    \ensuremath{\nu}=0.3. Fix z=0 face, prescribe uz=1e-4 m at z=L. Output reaction force and compare
    average \ensuremath{\sigma}zz to E*(\ensuremath{\Delta}L/L).
  \item Solid mechanics with thermal expansion (easy): 1D bar (L=1 m), E=200 GPa, \ensuremath{\nu}=0.3,
    \ensuremath{\alpha}=12e-6 1/K. Fix both ends (u=0 at x=0 and x=L). Apply uniform
    \ensuremath{\Delta}T=100 K via a temperature field. Output axial stress and reaction forces.
  \item Thermal expansion (2D, simple constraints): 2D plane stress plate (1\ensuremath{\times}1 m),
    E=100 GPa, \ensuremath{\nu}=0.3, \ensuremath{\alpha}=10e-6 1/K. Pin the bottom-left corner (ux=uy=0)
    and constrain bottom-right corner uy=0 (to prevent rigid motion but allow expansion). Apply uniform
    \ensuremath{\Delta}T=50 K. Output displacement at the top-right corner.
  \item Create and run a 3D linear elastic block (0.1\ensuremath{\times}0.02\ensuremath{\times}0.02 m),
    E=210 GPa, \ensuremath{\nu}=0.3. Fix x=0 face (ux=uy=uz=0). Prescribe ux=1e-4 m on x=L face (uy, uz
    free). Output reaction force on x=L and \ensuremath{\sigma}xx field.
  \item Create and run a 3D simple shear on a cube (0.05 m side), E=50 GPa, \ensuremath{\nu}=0.3. Fix
    the bottom face. Prescribe a uniform x-displacement of 5e-5 m on the top face (and set uz=0 there),
    with minimal constraints to avoid rigid motion. Output \ensuremath{\tau}xy at the cube center.
  \item Do a simple mesh refinement check (still linear elastic): 2D plane-stress cantilever (L=1 m,
    H=0.1 m), E=200 GPa, \ensuremath{\nu}=0.3, uniform downward traction on right edge 1 MPa. Run once
    on a coarse mesh and once on a finer mesh and report the percent difference in tip uy.
  \item Create a minimal 2D linear elastic model that would have rigid body modes if unconstrained: a
    1\ensuremath{\times}1 m square, E=10 GPa, \ensuremath{\nu}=0.3, with a small uniform body force in
    x. Add the smallest set of constraints to eliminate rigid motion (pin one node in x and y, plus one
    more constraint if needed). Run and output reactions to confirm equilibrium.
  \item Build and run a 2D plane-strain rectangle (1\ensuremath{\times}0.5 m), E=30 GPa,
    \ensuremath{\nu}=0.25. Fix left edge (ux=uy=0). Apply uniform horizontal traction 2 MPa on the right
    edge and uniform downward traction 1 MPa on the top edge. Output total reaction forces on the left
    edge and bottom edge (should balance applied loads).
\end{enumerate}}

\subsection{Porous flow}
{\small\sloppy\setlength{\emergencystretch}{3em}%
\begin{enumerate}[label=P\arabic*., leftmargin=*, align=left, itemsep=0.3em, topsep=0.3em]
  \item Create and run a steady 1D Darcy flow in a homogeneous column (L=1 m). Use constant permeability
    k=1e-12 m\textsuperscript{2} and viscosity mu=1e-3 Pa\ensuremath{\cdot}s (ignore gravity). Set p=2e5
    Pa at x=0 and p=1e5 Pa at x=1. Output pressure vs x and the Darcy flux.
  \item Create and run a steady 1D Darcy flow (L=1 m) with p=1e5 Pa at x=0 and p=0 at x=1. Use k=1e-12
    m\textsuperscript{2}, mu=1e-3 Pa\ensuremath{\cdot}s. Output the flux and confirm it is constant
    along the domain.
  \item Create and run a transient 1D slightly compressible flow (pressure diffusion) in a porous bar
    (L=1 m). Use k=1e-12 m\textsuperscript{2}, porosity=0.2, mu=1e-3 Pa\ensuremath{\cdot}s, ct=1e-9
    1/Pa. Initial p=1e5 Pa. At t=0 set x=0 to p=2e5 Pa and keep x=1 at no-flow. Run 0--2000 s and output
    p(t) at x=0.5 m.
  \item Create and run a transient 1D pressure diffusion case (L=1 m) with both ends fixed pressure:
    p=2e5 Pa at x=0 and p=1e5 Pa at x=1. Use k=1e-12, porosity=0.2, mu=1e-3, ct=1e-9. Initial p=1e5 Pa.
    Run 0--1000 s and output p(t) at x=0.5.
  \item Build and run a steady 2D Darcy flow on a unit square (1\ensuremath{\times}1). Use k=1e-12
    m\textsuperscript{2}, mu=1e-3 Pa\ensuremath{\cdot}s. Set p=2e5 Pa on the left boundary and p=1e5 Pa
    on the right boundary. Set no-flow on top/bottom. Output pressure contours and velocity vectors.
  \item Build and run a steady 2D Darcy flow on a 2\ensuremath{\times}1 rectangle. Use k=1e-12, mu=1e-3.
    Left boundary p=2e5 Pa, right boundary p=1e5 Pa, top/bottom no-flow. Output the integrated outflow
    on the right boundary.
  \item Create and run a steady 2D Darcy flow on a unit square with a two-zone permeability: left half
    k=1e-12, right half k=1e-13 (mu=1e-3). Set left boundary p=2e5, right boundary p=1e5, top/bottom
    no-flow. Output pressure and show the gradient changes across the interface.
  \item Create and run a steady 1D two-layer Darcy flow (L=1 m): k=1e-12 for x<0.5 and k=1e-13 for
    x>=0.5 (mu=1e-3). Set p=2e5 at x=0 and p=1e5 at x=1. Output p(x) and confirm the flux is constant.
  \item Create and run a steady 2D Darcy flow with anisotropic permeability: kx=1e-12, ky=1e-13
    (mu=1e-3). Domain 1\ensuremath{\times}1. Left p=2e5, right p=1e5, top/bottom no-flow. Output vx and
    vy fields.
  \item Create and run a steady 2D Darcy flow on a unit square with a high-permeability stripe:
    background k=1e-13, stripe k=1e-12 (mu=1e-3). Left p=2e5, right p=1e5, top/bottom no-flow. Output
    velocity magnitude to show channeling.
  \item Create and run a transient 2D pressure diffusion case on a unit square. Use k=1e-12,
    porosity=0.2, mu=1e-3, ct=1e-9. Initial p=1e5. Set left boundary to p=2e5 at t=0, right boundary
    p=1e5, and top/bottom no-flow. Run 0--1000 s and output p(t) at the center.
  \item Create and run a transient 2D pressure diffusion case on a 1\ensuremath{\times}1 square with all
    boundaries no-flow. Initial p=1e5. Add a constant uniform volumetric source term for 0<t<10 s then
    turn it off. Use k=1e-12, porosity=0.2, mu=1e-3, ct=1e-9. Run to 100 s and output average pressure
    vs time.
  \item Create and run a transient 1D closed-domain storage test (L=1 m, no-flow both ends). Use
    porosity=0.2, ct=1e-9, k=1e-12, mu=1e-3. Initial p=1e5. Apply a uniform source for 0<t<5 s then a
    uniform sink of equal magnitude for 5<t<10 s. Output average pressure vs time (should return near
    the initial value).
  \item Create and run a steady 1D Darcy flow with a prescribed constant flux at x=0 and fixed pressure
    at x=1. Choose a simple flux value (e.g., q=1e-6 m/s) and p=1e5 Pa at x=1. Use k=1e-12, mu=1e-3.
    Output p(x) and confirm the boundary flux is matched.
  \item Create and run a steady 2D Darcy flow with mixed BCs: left boundary fixed pressure p=2e5, right
    boundary no-flow, top boundary fixed pressure p=1e5, bottom no-flow. Domain 1\ensuremath{\times}1,
    k=1e-12, mu=1e-3. Output pressure contours and velocity vectors.
  \item Create and run a simple gravity-on hydrostatic test in 1D (vertical column height=1 m). Use
    rho=1000 kg/m\textsuperscript{3} and gravity. Set pressure at the top to 1e5 Pa and initialize with
    hydrostatic pressure. Run a short transient and report max velocity (should be near zero).
  \item Create and run a steady 2D Darcy flow with gravity turned off, then turn gravity on and keep the
    same left/right pressure BCs. Domain 1\ensuremath{\times}1, rho=1000, mu=1e-3, k=1e-12. Run both
    cases and output the velocity field to confirm gravity changes the solution only if vertical
    gradients/BCs exist.
  \item Create and run a manufactured-solution verification on a 2D unit square (steady). Choose
    p(x,y)=sin(pi*x)*sin(pi*y). Use constant k and mu, add the matching source term and Dirichlet BCs
    from the exact solution. Compute L2 error.
  \item Create and run a manufactured-solution verification on a 1D domain (steady). Choose
    p(x)=sin(pi*x) on x in [0,1]. Use constant k and mu, add the matching source term and Dirichlet BCs.
    Compute L2 error.
  \item Create and run a transient 1D diffusion with a time-dependent left boundary pressure: p(0,t)=1e5
    + 5e4*sin(2*pi*t/100) Pa, and no-flow at x=1. Use k=1e-12, porosity=0.2, mu=1e-3, ct=1e-9. Run
    0--500 s and output p(t) at x=0.5.
  \item Create and run a transient 1D diffusion with a step boundary condition: initial p=1e5
    everywhere, then set p=2e5 at x=0 for t>0 while x=1 stays at p=1e5. Use k=1e-12, porosity=0.2,
    mu=1e-3, ct=1e-9. Output p(t) at x=0.25 and x=0.75.
  \item Create and run a steady 2D Darcy flow on a coarse mesh (unit square). Use k=1e-12, mu=1e-3, left
    p=2e5, right p=1e5, top/bottom no-flow. Keep the mesh very coarse (e.g., 10\ensuremath{\times}10) to
    ensure quick runtime. Output pressure and velocity.
  \item Create and run a steady 2D Darcy flow and add a couple of point probes: domain
    1\ensuremath{\times}1, k=1e-12, mu=1e-3, left p=2e5, right p=1e5, top/bottom no-flow. Output p(t)
    (steady so constant) or p(step) at (0.25,0.5) and (0.75,0.5) to CSV.
  \item Create and run a steady 1D Darcy flow with units kept simple and minimal parameters: set k/mu =
    1 (use any consistent values), domain [0,1], p(0)=1, p(1)=0. Output p(x) and flux.
  \item Create and run a steady 2D Darcy flow with a low-permeability block (simple geometry): unit
    square with a centered 0.2\ensuremath{\times}0.2 block having k=1e-14 while background k=1e-12
    (mu=1e-3). Left p=2e5, right p=1e5, top/bottom no-flow. Output pressure and velocity magnitude.
\end{enumerate}}

\subsection{Navier--Stokes}
{\small\sloppy\setlength{\emergencystretch}{3em}%
\begin{enumerate}[label=N\arabic*., leftmargin=*, align=left, itemsep=0.3em, topsep=0.3em]
  \item Create a minimal 2D steady incompressible channel flow (Poiseuille) on [0,2]x[0,1]. Use
    prescribed inlet velocity profile, no-slip walls, and pressure outlet. Run and output velocity and
    pressure.
  \item Write a 2D transient start-up channel flow: start from u=v=0, ramp inlet velocity from 0 to U
    over t=[0,1]. Run and output centerline velocity vs time.
  \item Generate a 2D lid-driven cavity flow on a unit square: no-slip walls, moving lid at top with
    u=1. Run and output velocity magnitude.
  \item Create a 2D Couette flow: top wall moving, bottom stationary, periodic left/right,
    incompressible solve. Run and compare with linear velocity profile.
  \item Build a 2D steady Stokes flow test (very low Re): drive with a constant body force in x, no-slip
    top/bottom, periodic in x. Run and output u(y).
  \item Create a 2D Taylor--Green vortex decay in a periodic box by using an initial velocity field. Run
    transient and output kinetic energy decay vs time.
  \item Make a 2D channel flow case with symmetry: model only half the channel with a symmetry boundary
    at the centerline. Run and output profile.
  \item Create a 2D pressure-driven channel flow using pressure BCs (p\_in and p\_out) instead of a
    velocity inlet. Run and output the resulting flow rate.
  \item Generate a 3D steady Poiseuille-like flow in a short 3D box channel (coarse mesh): no-slip on
    walls, pressure difference from inlet to outlet. Run it.
  \item Create a 2D open-channel style rectangle with left inlet velocity, right outlet pressure, and
    no-slip top/bottom. Run and output pressure field.
  \item Write a 2D incompressible flow input using a finite-volume Navier--Stokes formulation (if
    available in your build). Run it on a coarse mesh.
  \item Write a 2D incompressible flow input using a finite-element Navier--Stokes formulation (if
    available). Run it on a coarse mesh.
  \item Create a 2D transient incompressible flow case that uses adaptive time stepping to maintain
    stability (e.g., based on nonlinear iterations). Run it.
  \item Build a 2D channel flow with variable viscosity mu(y)=mu0*(1+y) via a Material. Run and output
    u(y).
  \item Create a 2D steady flow that outputs volumetric flow rate at the outlet using a Postprocessor
    and writes it to CSV. Run it.
  \item Generate a 2D channel case and compute inlet vs outlet flow rate to check mass conservation
    (Postprocessors). Run and output both to CSV.
  \item Create a lid-driven cavity case and output the velocity at the cavity center vs time to CSV. Run
    it.
  \item Make a 2D channel flow case with a time-dependent inlet velocity (e.g., u\_in=sin(t) clipped to
    positive). Run a short transient.
  \item Create a 2D channel flow with a very coarse mesh (nx\ensuremath{\sim}12, ny\ensuremath{\sim}6)
    that runs fast. Solve steady and run it.
  \item Generate a 3D lid-driven cavity flow (coarse mesh) and run a short transient; output velocity
    magnitude.
  \item Create an incompressible flow case that writes Exodus plus a CSV of pressure drop between two
    points. Run it.
  \item Create a 2D transient channel flow with a sudden inlet velocity step (u\_in jumps from 0 to 1 at
    t=0). Run to steady state and output velocity magnitude vs time.
  \item Create a 2D steady flow case using PETSc defaults (no fancy solver tuning) and ensure it runs.
  \item Generate a 2D channel flow case that outputs the L2 norm of the divergence (or continuity
    residual) as a diagnostic Postprocessor. Run it.
  \item Create a minimal incompressible Navier--Stokes input expected to work with your NavierStokes
    module app (mesh, velocity/pressure variables, BCs, Executioner). Run it.
\end{enumerate}}

\subsection{Phase field}
{\small\sloppy\setlength{\emergencystretch}{3em}%
\begin{enumerate}[label=PF\arabic*., leftmargin=*, align=left, itemsep=0.3em, topsep=0.3em]
  \item Create a minimal 2D Cahn--Hilliard spinodal decomposition on a unit square with periodic BCs and
    a small random-noise initial condition about c=0.5. Run and output concentration to Exodus.
  \item Write a 2D Cahn--Hilliard problem with a Gaussian initial concentration blob that
    coarsens/relaxes. Run transient and output the field.
  \item Generate a 1D Cahn--Hilliard interface relaxation test starting from a tanh profile. Run and
    output c(x,t) snapshots.
  \item Create a 2D Cahn--Hilliard case with two initial droplets in a matrix and run coarsening; output
    concentration and free energy.
  \item Build a 2D Allen--Cahn shrinking-circle test: initialize a circular nucleus of phase A in phase
    B. Run and track the nucleus area vs time.
  \item Write a 1D Allen--Cahn interface-motion test with a double-well potential. Run transient and
    output the order parameter profile.
  \item Create a 2D Allen--Cahn coarsening test with random initial order parameter and periodic BCs.
    Run and output the evolution.
  \item Make a 2D Allen--Cahn case with spatially varying mobility (e.g., M(x)=1+0.5*x via Material).
    Run and compare interface speeds.
  \item Create a 2D Cahn--Hilliard case with concentration-dependent mobility provided by a Material.
    Run and output c and mobility.
  \item Generate a 2D spinodal decomposition case that uses adaptive time stepping to keep nonlinear
    solves stable. Run it.
  \item Produce a small 3D Cahn--Hilliard spinodal decomposition on a coarse cube mesh (fast). Run a
    short transient and output to Exodus.
  \item Create a 2D grain-growth style phase-field setup using multiple order parameters (3--5 grains)
    and run a short coarsening simulation.
  \item Generate a 2D multi-order-parameter grain growth case with \ensuremath{\sim}10 seeded grains and
    periodic BCs. Run and output grain IDs (or order parameters).
  \item Create a 2D phase-field case with a pinned region: set mobility \ensuremath{\sim}0 in a square
    subregion to act like an obstacle. Run and show interface pinning.
  \item Build a 2D Allen--Cahn case with a time-dependent driving force (tilt the double well with a
    Function). Run and output the mean order parameter vs time.
  \item Create a 2D Cahn--Hilliard case with zero-flux (natural) boundaries on all sides (no periodic).
    Run and output concentration evolution.
  \item Write a 2D Cahn--Hilliard case with Dirichlet concentration on left/right boundaries to enforce
    a gradient, and natural/no-flux on top/bottom. Run it.
  \item Create a 2D Allen--Cahn case with periodic BCs and output the domain-averaged order parameter vs
    time to CSV. Run it.
  \item Generate a phase-field input that computes total free energy vs time using a Postprocessor and
    writes it to CSV. Run it.
  \item Make an Allen--Cahn case that outputs an approximate interface measure using an integral of
    |grad(eta)| (or similar) and writes to CSV. Run it.
  \item Create a 2D Cahn--Hilliard simulation using SECOND-order Lagrange elements and run it.
  \item Produce a tiny phase-field test case that runs fast (coarse mesh, short end time) but still
    evolves. Run it.
  \item Create a 2D Allen--Cahn phase-field case with two initial circular inclusions that merge over
    time. Run a short transient and output the phase-field variable to Exodus.
  \item Write a 2D Cahn--Hilliard case where the initial condition is defined by a ParsedFunction (e.g.,
    sinusoidal pattern). Run it.
  \item Create a minimal PhaseField-module input expected to work in common MOOSE module apps
    (GeneratedMesh, variable[s], phase-field kernels/actions, Executioner). Run it.
\end{enumerate}}

\subsection{\(J_2\) plasticity}
{\small\sloppy\setlength{\emergencystretch}{3em}%
\begin{enumerate}[label=J\arabic*., leftmargin=*, align=left, itemsep=0.3em, topsep=0.3em]
  \item Make a 1D bar (L=1) with J2 plasticity (E=200e9, nu=0.3, sigma\_y=200e6, H=1e9). Fix left end,
    pull right end to 1\% strain. Run and output stress vs strain.
  \item 1D bar J2 perfect plasticity (E=200e9, nu=0.3, sigma\_y=200e6, H=0). Fix left end, pull right end to 1\%
    strain. Output stress--strain.
  \item 1D bar J2 plasticity (E=100e9, nu=0.3, sigma\_y=100e6, H=5e8). Pull to 0.5\% strain. Output
    stress--strain and plastic strain.
  \item 1D bar J2 plasticity (E=50e9, nu=0.3, sigma\_y=50e6, H=1e8). Pull to 1\% strain. Output
    stress--strain.
  \item 1D bar J2 plasticity (E=10e9, nu=0.3, sigma\_y=5e6, H=1e7). Pull to 1\% strain. Output
    stress--strain.
  \item 1D bar J2 plasticity (E=200e9, nu=0.3, sigma\_y=300e6, H=1e9). Pull to 1\% strain. Output
    stress--strain.
  \item 1D bar J2 plasticity (E=200e9, nu=0.3, sigma\_y=150e6, H=2e9). Pull to 1\% strain. Output
    stress--strain.
  \item 1D bar J2 plasticity (E=200e9, nu=0.3, sigma\_y=150e6, H=1e8). Pull to 1\% strain. Output
    stress--strain.
  \item 1D bar J2 plasticity (E=200e9, nu=0.3, sigma\_y=150e6, H=0). Pull to 2\% strain. Output
    stress--strain.
  \item 1D bar J2 plasticity (E=70e9, nu=0.33, sigma\_y=120e6, H=3e8). Pull to 1\% strain. Output
    stress--strain.
  \item 1D bar J2 plasticity (E=200e9, nu=0.3, sigma\_y=200e6, H=1e9). Fix left end, compress right end
    to -1\% strain. Output stress--strain.
  \item 1D bar J2 perfect plasticity (E=200e9, nu=0.3, sigma\_y=200e6, H=0). Compress to -1\% strain.
    Output stress--strain.
  \item 1D bar J2 plasticity (E=100e9, nu=0.3, sigma\_y=100e6, H=5e8). Compress to -0.5\% strain. Output
    stress--strain.
  \item 1D bar J2 plasticity (E=50e9, nu=0.3, sigma\_y=50e6, H=1e8). Compress to -1\% strain. Output
    stress--strain.
  \item 1D bar J2 plasticity (E=10e9, nu=0.3, sigma\_y=5e6, H=1e7). Compress to -1\% strain. Output
    stress--strain.
  \item 1D bar load-unload: J2 plasticity (E=200e9, nu=0.3, sigma\_y=200e6, H=1e9). Pull to 1\% strain
    then unload back to 0. Output stress--strain (show residual strain).
  \item 1D bar load-unload: J2 perfect plasticity (E=200e9, nu=0.3, sigma\_y=200e6, H=0). Pull to 1\%
    then unload to 0. Output stress--strain.
  \item 1D bar load-unload: J2 plasticity (E=100e9, nu=0.3, sigma\_y=100e6, H=5e8). Pull to 0.5\% then
    unload to 0. Output stress--strain.
  \item 1D bar load-unload: J2 plasticity (E=50e9, nu=0.3, sigma\_y=50e6, H=1e8). Pull to 1\% then
    unload to 0. Output stress--strain.
  \item 1D bar load-unload: J2 plasticity (E=10e9, nu=0.3, sigma\_y=5e6, H=1e7). Pull to 1\% then unload
    to 0. Output stress--strain.
  \item 1D bar traction control: J2 plasticity (E=200e9, nu=0.3, sigma\_y=200e6, H=1e9). Fix left end,
    ramp traction on right end from 0 to 300 MPa. Output strain vs applied traction.
  \item 1D bar traction control: J2 perfect plasticity (E=200e9, nu=0.3, sigma\_y=200e6, H=0). Ramp
    traction 0 to 300 MPa. Output strain vs traction.
  \item 1D bar traction control (elastic-only check): J2 plasticity (E=200e9, nu=0.3, sigma\_y=400e6,
    H=1e9). Ramp traction 0 to 100 MPa (below yield). Output plastic strain (should stay
    \ensuremath{\sim}0).
  \item 1D bar traction control (yield check): J2 plasticity (E=200e9, nu=0.3, sigma\_y=120e6, H=5e8).
    Ramp traction 0 to 200 MPa. Output when plastic strain first becomes nonzero.
  \item 1D bar traction control (soft material): J2 plasticity (E=10e9, nu=0.3, sigma\_y=5e6, H=1e7).
    Ramp traction 0 to 20 MPa. Output strain vs traction.
\end{enumerate}}

\subsection{Reactor module (mesh workflows)}
{\small\sloppy\setlength{\emergencystretch}{3em}%
\begin{enumerate}[label=R\arabic*., leftmargin=*, align=left, itemsep=0.3em, topsep=0.3em]
  \item Create a minimal hex ReactorMeshParams setup with a single pin mesh. Run it and output the mesh
    to Exodus.
  \item Build a pin-cell mesh with fuel, gap, and clad rings plus named regions. Run it and output a
    mesh summary with region names.
  \item Generate a simple hex assembly from repeated pin units. Run it and output block IDs to verify
    assembly construction.
  \item Create a small core mesh using one assembly type in a compact pattern. Run it and output the
    full core mesh.
  \item Build a core mesh with dummy\_assemblies at selected positions. Run it and output a region map
    showing empty locations.
  \item Add one PeripheralRingMeshGenerator layer around a core mesh. Run it and output boundary names
    and IDs.
  \item Create a basic Cartesian pin mesh in an HP-MR style workflow. Run it and output the generated
    mesh.
  \item Apply RenameBoundaryGenerator to a reactor mesh and rename at least two boundaries. Run it and
    print the renamed boundary list.
  \item Use AdvancedExtruderGenerator to extrude a 2D reactor mesh into 3D. Run it and output the 3D
    mesh to Exodus.
  \item Add PlaneIDMeshGenerator to a reactor mesh workflow. Run it and output plane ID and region ID
    information.
  \item Perform a subdomain ID mapping validation at pin level for fuel, clad, and coolant IDs. Run it
    and report expected-vs-actual subdomain IDs in a table.
  \item Generate a basic ABTR-style pin-lattice assembly mesh. Run it and output assembly ID labeling in
    the results.
  \item Create a reactor mesh and generate side sets from boundary normals for later BC use. Run it and
    output the side set names.
  \item Build a small core with two assembly IDs in a simple pattern. Run it and output assembly
    distribution by ID.
  \item Enable RGMB-related reporting metadata output for a simple core mesh. Run it and capture an ID
    summary table in console output.
  \item Create an ABTR-like assembly mesh with ring intervals and ducts. Run it and output a block map
    with IDs.
  \item Build an ABTR-like multi-assembly core pattern with a peripheral ring. Run it and output
    assembly and region IDs.
  \item Use HexagonConcentricCircleAdaptiveBoundaryMeshGenerator around an assembly mesh. Run it and
    output interface/boundary quality information.
  \item Create a mixed hex and Cartesian reactor mesh composition workflow. Run it and output merged
    mesh block and region IDs.
  \item Apply multiple mesh operations in sequence (rename boundaries plus delete/keep selected
    boundaries). Run it and output the final boundary set.
  \item Perform a subdomain ID mapping validation at assembly level across multiple pin types. Run it
    and report expected-vs-actual subdomain IDs with mismatch checks.
  \item Perform a subdomain ID mapping validation at core level across assembly IDs and peripheral
    region IDs. Run it and output a full expected-vs-actual mapping table.
  \item Build an intermediate RGMB workflow combining control-drum structures with assemblies in one
    core. Run it and output reporting IDs for each structure.
  \item Create a conformal component stitching workflow for reactor mesh components. Run it and output
    stitched interface IDs and block IDs.
  \item Generate a Cartesian core-like HP-MR style pattern with multiple regions. Run it and output a
    subdomain and boundary ID summary.
\end{enumerate}}

\section{Methods of Manufactured Solutions Benchmark Problem Definitions}
\label{app:mms-problems}

The Methods of Manufactured Solutions (MMS) benchmark evaluates whether \gls{moosenger} can construct MOOSE inputs that are not only syntactically valid and semantically aligned with a prompt, but also reproduce a known analytical solution, which would then be used for physics verification. MMS is well suited to this purpose because the exact solution is prescribed before the numerical problem is constructed. The corresponding source terms, boundary conditions, and initial conditions are then derived so that the prescribed field satisfies the governing equations exactly.

Let $\phi^{*}$ denote a manufactured analytical solution and let $\mathcal{L}$ denote the differential operator of the selected physical model. The manufactured source is obtained by applying the governing operator to the exact solution:
\begin{equation}
    s^{*} = \mathcal{L}\left(\phi^{*}\right).
\end{equation}
The numerical solution $\phi_h$ produced by MOOSE can therefore be compared directly with $\phi^{*}$. This approach tests the complete input-generation
workflow, including the selection of MOOSE objects, construction of source functions, assignment of material properties, and enforcement of boundary and initial conditions, as well as the solver configuration, execution, and extraction of verification quantities.

The benchmark contains ten problems spanning steady and transient diffusion, heat conduction, phase-field evolution, porous-media flow, solid mechanics, fluid flow, and coupled thermoelasticity. All problems are defined on the 2D unit-square domain:
\begin{equation}
    \Omega = (0,1)\times(0,1).
\end{equation}

Table~\ref{tab:mms-cases} summarizes the governing equations,
manufactured solutions, source terms, and boundary and initial conditions
for the ten MMS problems. Here, $\mathbf{d}=(d_x,d_y)$ denotes displacement
and $\mathbf{u}=(u,v)$ denotes velocity. The small-strain tensor is:
\begin{equation}
    \boldsymbol{\varepsilon}(\mathbf{d})
    =
    \frac{1}{2}
    \left(
        \nabla\mathbf{d}
        +
        \nabla\mathbf{d}^{T}
    \right).
\end{equation}

\newcolumntype{L}[1]{>{\RaggedRight\arraybackslash}p{#1}}
{\footnotesize
\setlength{\tabcolsep}{6pt}
\renewcommand{\arraystretch}{1.15}

{\footnotesize

\setlength{\tabcolsep}{5pt}
\setlength{\LTleft}{0pt}
\setlength{\LTright}{0pt}
\renewcommand{\arraystretch}{1.15}

\begin{longtable}{
@{}
L{0.63\textwidth}
L{0.31\textwidth}
@{}
}

\caption{The ten MMS benchmark problems. Source terms are obtained by
substituting the manufactured solution into the corresponding governing
equation.}
\label{tab:mms-cases}\\

\toprule
\textbf{Governing model and manufactured solution} &
\textbf{Boundary and initial conditions} \\
\midrule
\endfirsthead

\multicolumn{2}{c}{
\tablename~\thetable{} (continued)
}\\[2pt]

\toprule
\textbf{Governing model and manufactured solution} &
\textbf{Boundary and initial conditions} \\
\midrule
\endhead

\midrule
\multicolumn{2}{r}{
\textit{Continued on next page}
}\\
\endfoot

\bottomrule
\endlastfoot

\multicolumn{2}{@{}l@{}}{
\textbf{1. Steady diffusion---Dirichlet}
}\\[-2pt]
\cmidrule(lr){1-2}

The governing equation is
\[
-\nabla^{2}u=f,
\qquad
D=1.
\]

The manufactured solution is
\[
u^{*}
=
\sin(\pi x)\sin(\pi y),
\]
which gives
\[
f
=
2\pi^{2}
\sin(\pi x)\sin(\pi y).
\]
&
\textbf{BC:}
\[
u=0
\qquad
\text{on }\partial\Omega .
\]

\textbf{IC:} None.
\\
\addlinespace[6pt]

\multicolumn{2}{@{}l@{}}{
\textbf{2. Steady diffusion---Neumann}
}\\[-2pt]
\cmidrule(lr){1-2}

The governing equation is
\[
-\nabla^{2}u=f,
\qquad
D=1,
\]
with the domain-average constraint
\[
\langle u\rangle_{\Omega}=2.
\]

The manufactured solution is
\[
u^{*}
=
2+\cos(\pi x)\cos(\pi y),
\]
which gives
\[
f
=
2\pi^{2}
\cos(\pi x)\cos(\pi y).
\]
&
\textbf{BC:}
\[
\partial_n u
=
\nabla u\cdot\mathbf{n}
=
0
\]
on \(\partial\Omega\).

The mean constraint
\(\langle u\rangle_{\Omega}=2\)
removes the constant nullspace.

No Dirichlet condition is applied.

\textbf{IC:} None.
\\
\addlinespace[6pt]

\multicolumn{2}{@{}l@{}}{
\textbf{3. Transient diffusion---Dirichlet}
}\\[-2pt]
\cmidrule(lr){1-2}

The governing equation is
\[
\frac{\partial u}{\partial t}
-
\nabla^{2}u
=
s,
\]
with
\[
D=1,
\qquad
t_f=1.
\]

The manufactured solution is
\[
u^{*}
=
(1+t)
\sin(\pi x)\sin(\pi y),
\]
which gives
\[
s
=
\left[
1+2\pi^{2}(1+t)
\right]
\sin(\pi x)\sin(\pi y).
\]
&
\textbf{BC:}
\[
u=0
\qquad
\text{on }\partial\Omega .
\]

\textbf{IC:}
\[
u(x,y,0)
=
\sin(\pi x)\sin(\pi y).
\]
\\
\addlinespace[6pt]

\multicolumn{2}{@{}l@{}}{
\textbf{4. Transient diffusion---Neumann}
}\\[-2pt]
\cmidrule(lr){1-2}

The governing equation is
\[
\frac{\partial u}{\partial t}
-
\nabla^{2}u
=
0,
\]
with
\[
D=1,
\qquad
t_f=0.05.
\]

The manufactured solution is
\[
u^{*}
=
2+
e^{-2\pi^{2}t}
\cos(\pi x)\cos(\pi y).
\]
&
\textbf{BC:}
\[
\partial_n u=0
\qquad
\text{on }\partial\Omega .
\]

\textbf{IC:}
\[
u(x,y,0)
=
2+\cos(\pi x)\cos(\pi y).
\]

The initial condition fixes the conserved mean, so an additional mean
constraint is not required.
\\
\addlinespace[6pt]

\multicolumn{2}{@{}l@{}}{
\textbf{5. Allen--Cahn phase field}
}\\[-2pt]
\cmidrule(lr){1-2}

The governing equation is
\[
\frac{\partial\eta}{\partial t}
=
M
\left[
\kappa\nabla^{2}\eta
-
F'(\eta)
\right]
+
S,
\]
with
\[
M=1,
\qquad
\kappa=0.01,
\]
and
\[
F(\eta)
=
\eta^{2}(1-\eta)^{2}.
\]

The manufactured solution is
\[
\eta^{*}
=
\frac{1}{2}
+
\frac{1}{10}
e^{-t}
\sin(\pi x)\sin(\pi y).
\]

The corresponding source is
\[
S
=
\partial_t\eta^{*}
-
M
\left[
\kappa\nabla^{2}\eta^{*}
-
F'(\eta^{*})
\right].
\]

The evaluated input used \(t_f=1\).
&
\textbf{BC:}
\[
\eta=\eta^{*}
\qquad
\text{on }\partial\Omega .
\]

\textbf{IC:}
\[
\eta(x,y,0)
=
\frac{1}{2}
+
\frac{1}{10}
\sin(\pi x)\sin(\pi y).
\]
\\
\addlinespace[6pt]

\multicolumn{2}{@{}l@{}}{
\textbf{6. Single-phase Darcy pressure}
}\\[-2pt]
\cmidrule(lr){1-2}

The governing equation is
\[
-\nabla\cdot
\left(
\frac{\kappa}{\mu}\nabla p
\right)
=
q,
\]
with
\[
\mu=1,
\qquad
\kappa(x)=1+\frac{x}{4}.
\]

The manufactured pressure is
\[
p^{*}
=
1+\sin(\pi x)\sin(\pi y).
\]

The corresponding source is
\[
\begin{aligned}
q
={}&
\frac{\pi}{4}
\Big[
2\pi(x+4)\sin(\pi x)
\\
&\qquad
-\cos(\pi x)
\Big]
\sin(\pi y).
\end{aligned}
\]
&
\textbf{BC:}
\[
p=p^{*}
\qquad
\text{on }\partial\Omega .
\]

Because the sine terms vanish on the external boundaries,
\[
p=1
\qquad
\text{on }\partial\Omega .
\]

\textbf{IC:} None.
\\
\addlinespace[6pt]

\multicolumn{2}{@{}l@{}}{
\textbf{7. Transient heat conduction}
}\\[-2pt]
\cmidrule(lr){1-2}

The governing equation is
\[
\rho c_p
\frac{\partial T}{\partial t}
-
\nabla\cdot(k\nabla T)
=
Q,
\]
with
\[
\rho=c_p=k=1.
\]

The manufactured solution is
\[
T^{*}
=
e^{-t}
\sin(\pi x)\sin(\pi y),
\]
which gives
\[
Q
=
(2\pi^{2}-1)
e^{-t}
\sin(\pi x)\sin(\pi y).
\]

The evaluated input used \(t_f=1\).
&
\textbf{BC:}
\[
T=T^{*}
\qquad
\text{on }\partial\Omega .
\]

For this manufactured solution,
\[
T=0
\qquad
\text{on }\partial\Omega .
\]

\textbf{IC:}
\[
T(x,y,0)
=
\sin(\pi x)\sin(\pi y).
\]
\\
\addlinespace[6pt]

\multicolumn{2}{@{}l@{}}{
\textbf{8. Small-strain linear elasticity}
}\\[-2pt]
\cmidrule(lr){1-2}

The governing equation is
\[
-\nabla\cdot\boldsymbol{\sigma}
=
\mathbf{b},
\]
with
\[
\boldsymbol{\sigma}
=
\lambda
\operatorname{tr}
(\boldsymbol{\varepsilon})
\mathbf{I}
+
2\mu\boldsymbol{\varepsilon},
\]
and
\[
\lambda=2,
\qquad
\mu=3.
\]

The manufactured displacement is
\[
d_x^{*}
=
\sin(\pi x)\sin(\pi y),
\]
\[
d_y^{*}
=
\cos(\pi x)\sin(\pi y).
\]

The corresponding body force is
\[
\mathbf{b}
=
-\nabla\cdot
\boldsymbol{\sigma}
(\mathbf{d}^{*}).
\]
&
\textbf{BC:}

Exact Dirichlet displacement conditions are imposed on all external
boundaries:
\[
d_x=d_x^{*},
\qquad
d_y=d_y^{*}.
\]

\textbf{IC:} None.
\\
\addlinespace[6pt]

\multicolumn{2}{@{}l@{}}{
\textbf{9. Steady incompressible Navier--Stokes}
}\\[-2pt]
\cmidrule(lr){1-2}

The governing equations are
\[
\rho
(\mathbf{u}\cdot\nabla)\mathbf{u}
-
\mu\nabla^{2}\mathbf{u}
+
\nabla p
=
\mathbf{f},
\]
\[
\nabla\cdot\mathbf{u}=0,
\]
with
\[
\rho=\mu=1.
\]

The manufactured velocity and pressure fields are
\[
u^{*}
=
\sin^{2}(\pi x)\sin(2\pi y),
\]
\[
v^{*}
=
-\sin(2\pi x)\sin^{2}(\pi y),
\]
\[
p^{*}
=
\cos(\pi x)\cos(\pi y).
\]

The corresponding momentum source is
\[
\begin{aligned}
\mathbf{f}
={}&
(\mathbf{u}^{*}\cdot\nabla)\mathbf{u}^{*}
-
\nabla^{2}\mathbf{u}^{*}
\\
&+
\nabla p^{*}.
\end{aligned}
\]
&
\textbf{BC:}

Exact velocity Dirichlet conditions are imposed on
\(\partial\Omega\).

For the manufactured velocity,
\[
u=v=0
\qquad
\text{on }\partial\Omega .
\]

An exact pressure pin or equivalent pressure normalization is required to
remove the pressure nullspace.
\\
\addlinespace[6pt]

\multicolumn{2}{@{}l@{}}{
\textbf{10. Steady thermoelasticity}
}\\[-2pt]
\cmidrule(lr){1-2}

The governing equations are
\[
-\nabla^{2}T
=
Q,
\]
and
\[
-\nabla\cdot
\boldsymbol{\sigma}_{\mathrm{th}}
=
\mathbf{b}.
\]

The parameters are
\[
\lambda=2,
\qquad
\mu=3,
\qquad
\alpha=0.1,
\]
\[
k=1,
\qquad
T_0=0.
\]

The manufactured temperature and displacement fields are
\[
T^{*}
=
\sin(\pi x)\sin(\pi y),
\]
\[
d_x^{*}
=
\sin(\pi x)\sin(\pi y),
\]
\[
d_y^{*}
=
\cos(\pi x)\sin(\pi y).
\]

The thermal strain is
\[
\boldsymbol{\varepsilon}_{\mathrm{th}}
=
\alpha(T-T_0)\mathbf{I}.
\]

The thermal source is
\[
Q
=
-\nabla^{2}T^{*}
=
2\pi^{2}
\sin(\pi x)\sin(\pi y),
\]
and the mechanical body force is
\[
\mathbf{b}
=
-\nabla\cdot
\boldsymbol{\sigma}_{\mathrm{th}}
(T^{*},\mathbf{d}^{*}).
\]
&
\textbf{BC:}

Exact Dirichlet conditions are imposed for \(T\), \(d_x\), and \(d_y\)
on all external boundaries:
\[
T=T^{*},
\]
\[
d_x=d_x^{*},
\qquad
d_y=d_y^{*}.
\]

\textbf{IC:} None.
\\

\end{longtable}
}
}
\subsection{MMS Case Prompts}
\label{app:mms-prompt-template}

Every MMS case is presented to the agent as a single structured prompt. The
prompt states a fixed instruction header, followed by the complete case
specification: governing equation, geometry, variables, material constants,
prescribed analytical solution, pre-computed source terms, boundary and
initial conditions, evaluation quantities, and the exact output blocks to
include verbatim. Listing~\ref{lst:mms-prompt-dirichlet} shows the full prompt
for the steady Dirichlet diffusion case (the \texttt{explicit} variant used in
our reported runs). The analytical solution and forcing term are supplied
directly; the agent is instructed not to derive new forcing terms, so the task
is the faithful construction of a runnable MOOSE input rather than symbolic
manufacturing of the problem.

\begin{lstlisting}[style=mmsprompt,caption={Complete agent prompt for the steady
Dirichlet diffusion case. The manufactured solution and its compatible source
term are provided; the required \texttt{[Postprocessors]} and \texttt{[Outputs]}
blocks are included verbatim so that the numerical evaluator can read a uniform
\texttt{ElementL2Error} column.},label={lst:mms-prompt-dirichlet}]
Build an input-only MOOSE MMS verification input file from the full
specification below.

Use the provided source terms directly; do not derive new forcing terms.

Requirements:

- Produce a complete MOOSE input file.
- Use only input-file objects; do not require custom C++ code.
- Use exact Dirichlet boundary conditions unless the case notes say otherwise.
- Include the supplied L2-error postprocessors and CSV output block.
- Configure executioner, solver, mesh, variables, materials, kernels/physics,
  BCs, functions, and outputs.
- Run the completed input file using `mcp.run_input`.

Case specification:

- Case ID: diffusion_steady_dirichlet_2d
- Module: MOOSE Framework scalar kernels
- Physics: Steady scalar diffusion with a volumetric source and homogeneous
  Dirichlet boundaries
- Dimension: 2D unit square
- Time type: steady
- Equation: -div(grad(u)) = f

Variables:

- u

Constants:

- D = 1

Analytical solution:

- u = sin(pi*x)*sin(pi*y)

Forcing/source terms:

- f = 2*pi^2*sin(pi*x)*sin(pi*y)

Boundary conditions:

Homogeneous Dirichlet condition u = 0 on all external boundaries.

Initial conditions:

Not applicable for the steady solve.

Evaluation quantities:

- L2 error of u
- optional spatial convergence rate for u

Notes:

- This is the baseline steady scalar-diffusion calibration case.

Required evaluation blocks:

Include the blocks below in the final input. If the input already has
[Postprocessors] or [Outputs] sections, merge these child objects into those
sections without changing their names, variables, functions, or output routing.

[Postprocessors]
  [l2_error_u]
    type = ElementL2Error
    variable = u
    function = 'sin(pi*x)*sin(pi*y)'
    execute_on = FINAL
    outputs = csv
  []
[]

[Outputs]
  [csv]
    type = CSV
    file_base = mms_results
    execute_on = FINAL
  []
[]

After completing the input, run the final version with `mcp.run_input`.
\end{lstlisting}

Cases that do not use Dirichlet data specify their boundary treatment and any
additional constraints explicitly in the same slots. Listing~\ref{lst:mms-prompt-neumann}
shows the corresponding sections for the pure-Neumann diffusion case: the source
integrates to zero (the pure-Neumann compatibility condition), homogeneous
natural boundaries are requested on all sides, and a domain-average constraint
removes the constant nullspace and pins the intended member of the solution
family. Selecting an input-only constraint that realizes this mean-value
condition---rather than substituting a Dirichlet condition---is part of the task.

\begin{lstlisting}[style=mmsprompt,caption={Boundary-condition, forcing, and
notes sections that distinguish the pure-Neumann diffusion case from the
Dirichlet baseline. All other slots follow the same structure as
Listing~\ref{lst:mms-prompt-dirichlet}.},label={lst:mms-prompt-neumann}]
- Equation: -div(grad(u)) = f, mean(u) = 2

Constants:

- D = 1
- prescribed_mean_u = 2

Analytical solution:

- u = cos(pi*x)*cos(pi*y) + 2

Forcing/source terms:

- f = 2*pi^2*cos(pi*x)*cos(pi*y)

Boundary conditions:

Homogeneous Neumann condition grad(u) dot n = 0 on all external boundaries.
Impose the domain-average normalization mean(u) = 2 to remove the constant
nullspace; do not impose a Dirichlet boundary condition.

Notes:

- Use homogeneous natural Neumann boundaries and do not add a Dirichlet
  boundary condition.
- The source integrates to zero over the unit square, satisfying the
  pure-Neumann compatibility condition.
- The mean-value constraint selects the manufactured solution from the family
  that differs by an arbitrary constant.
- The normalization can be imposed input-only with a scalar Lagrange
  multiplier, ScalarLagrangeMultiplier, and AverageValueConstraint.
\end{lstlisting}

\end{document}